\documentclass{llncs}

\usepackage{hyperref}
\usepackage{comment}
\usepackage{dsfont}
\usepackage{stmaryrd}
\usepackage{graphicx}
\usepackage{wrapfig}

\newcommand{\real}{\mathds{R}}

\newcommand{\nat}{\mathds{N}}
\newcommand{\sem}[1]{{\llbracket #1 \rrbracket}}
\newcommand{\card}[1]{\vert #1 \vert}

\newcommand{\set}[1]{\{ #1 \}}
\newcommand{\prop}{\texttt{Prop}}

\newcommand{\all}[2]{\forall #1 \, . \, #2}
\newcommand{\ex}[2]{\exists #1 \, . \, #2}
\newcommand{\dall}[2]{\bar{\forall} #1 \, . \, #2}
\newcommand{\dex}[2]{\bar{\exists} #1 \, . \, #2}
\newcommand{\mn}[2]{\textsf{M} #1 \, . \, #2}
\newcommand{\argmax}{\mathop{\textrm{argmax}}}
\newcommand{\abs}[2]{\lambda #1 \, . \, #2}
\newcommand{\ltrue}{\top}
\newcommand{\lfalse}{\bot}
\newcommand{\crisp}{\mathop{\textrm{crisp}}}

\begin{document}

\title{Probabilistic Approximate Logic \\ and its Implementation \\
	in the Logical Imagination Engine}

\author
{
	Mark-Oliver Stehr\inst{1} \and 
	Minyoung Kim \inst{1} \and \\ 
	Carolyn L. Talcott\inst{1} \and
	Merrill Knapp\inst{1} \and 
	Akos Vertes\inst{2} 
}

\institute
{
	SRI International, Menlo Park, CA 94025\and
	Dept. of Chemistry, George Washington Univ., Washington, DC 20052
}

\date{April 2019}

\maketitle

\begin{abstract}
	In spite of the rapidly increasing number of applications of machine learning in various domains, a principled and
	systematic approach to the incorporation of domain knowledge in the engineering process is still lacking and ad hoc solutions that are difficult to validate are still the norm in practice, which is of growing concern not only in mission-critical applications. While AI has a long history of developing logics for knowledge representation, reasoning, and verification, we believe that in spite of rapid advances in both fields (cognitive and symbolic AI) there is a fundamental mismatch of technologies and foundations that is preventing unified solutions to emerge.
	
	In this note, we introduce Probabilistic Approximate Logic (PALO) as a logic based on the notion of mean approximate
	probability to overcome conceptual and computational difficulties inherent to strictly probabilistic logics.
	The logic is approximate in several dimensions. Logical independence assumptions are used to obtain approximate
	probabilities, but by averaging over many instances of formulas a useful estimate of mean probability with
	known confidence can usually be obtained. To enable efficient computational inference,
	the logic has a continuous semantics that reflects only a subset of the structural properties of classical logic,
	but this imprecision can be partly compensated by richer theories obtained by classical inference or other means. 
	Computational inference, which refers to the construction of models and validation of logical properties, is based
	on Stochastic Gradient Descent (SGD) and Markov Chain Monte Carlo (MCMC) techniques and hence another dimension where approximations are involved. The propositional core of PALO is based on ingredients from Hajek's Product logic, \L{}ukasiewicz logic, and G\"odel logic, but a non-standard semantics of quantifiers and theories is the key to a probabilistic interpretation and a practical implementation. 
	
	We also present the Logical Imagination Engine (LIME), a prototypical implementation of PALO based on
	TensorFlow. Albeit not limited to the biological domain, we illustrate its operation in a quite substantial bioinformatics machine learning application concerned with network synthesis and analysis in a recent DARPA project.
	
\end{abstract}

\pagestyle{plain}

\section{Introduction}

Data and knowledge can be both seen as instances of uncertain information, even if the source of uncertainty may not be the same. A lot of research in machine learning is naturally adopting a data-centric view, while research in logics is primarily concerned with knowledge and its manipulation. Models and uncertainty are key concepts that lie at the intersection of both fields and hence might be good starting points to lay the foundations for a more unified treatment that could serve as a basis for a better theoretical and technological integration. In practice, applications of machine learning should benefit from systematic methods to incorporate domain knowledge into the system engineering process. Conversely, logics and their implementations (e.g., validation/verification systems) should benefit from models that are learned from potentially large amounts of data in the context of other domain knowledge.

At the intersection, we have the concept of uncertainty that already has a long tradition in mathematics, engineering, computer science, and especially in the field of Artificial Intelligence. Often uncertainty is considered a nuisance or a challenge and pure qualitative and quantitative models, formalisms, and systems are extended to deal with uncertainty, sometimes in an ad hoc fashion viewing uncertainty as an add-on feature and an obstacle to a formally clean and computationally efficient treatment. A different approach is to think of uncertainty as an opportunity. Models and systems can be inherently based on uncertainty (very much like biological systems) and take advantage of uncertainty for conceptual and computational benefits. This seems to be the direction in which most of the research in machine learning is heading, especially with the probabilistic/approximate modeling that underpins many advances in the foundations of deep learning. In this note we would also like to use it as a motivation for PALO, our \textit{Probabilistic Approximate Logic}, which we first put into context with some traditional and related work.

Building on the work of Bacchus \cite{Bacchus91}, Halpern \cite{Halpern89}  considered two classical probabilistic first-order logics and their combination. The first logic assumes that domain of each variable constitutes a probability distribution, while the second uses a semantics where formulas are  interpreted w.r.t. a set of worlds which is equipped with a probability distribution. PALO is not a classical logic but rather a soft logic with a continuous interpretation of propositions as real numbers in the interval $[0,1]$. Domains are not equipped with probability distributions, but propositional formulas are interpreted as approximate probabilities. Such probabilities are lifted to quantified formulas by averaging to compensate for their approximate nature.  Probabilities are further lifted to models of theories in a way that compensates for the natural uncertainty in the weight/relevance of the axioms. This last point is essential because as a substructural logic a soft logic is necessarily incomplete w.r.t. to a classical semantics, which is more intuitive for the user to reason with, and hence an important reference influencing the design of PALO and its use.

\paragraph{Structure of this Paper} After discussing some important related work in the following section, we introduce the syntax of the PALO core language in Section \ref{sect-palo} together with primarily two flavors of semantics (soft and classical logic interpretations) to support different types of inference. In Section \ref{sect-lime}, we discuss our
current prototype implementation of PALO in what we call the \textit{Logical Imagination Engine} (LIME). This prototype
will be used in Section \ref{sect-bio} to illustrate the application of PALO to a network synthesis problem in the bioinformatics domain. This is the problem that originally motivated the development of PALO and its implementation. A number of extensions and opportunities for future work in the context of PALO and LIME will then be discussed in 
Section \ref{sect-ext}, followed by a slightly broader view of a potential role of PALO for addressing some key limitations of deep  learning architectures in the conclusion.

\section{Related Work}

The idea of viewing probabilities as generalized truth values goes back to Reichenbach \cite{Reichenbach49} but has remained unsatisfactory due to the non-extensional (in other words, non-truth-functional or non-denotational) nature of this interpretation. Generally, the probability $p(\phi \land \psi)$ can not be defined as a function of $p(\phi)$ and $p(\psi)$, and assumptions need to be made to achieve a strict correspondence to a meaningful probability. Our approach is conceptually related to a solution proposed by Gaines \cite{Gaines78}. He defines what he calls a \textit{Standard Uncertainty Logic} (SUL) that by simple axiomatic extensions yields either \textit{Stochastic Logic} or \textit{Fuzzy Logic}. Both logics can be equipped with a population semantics (not necessarily limited to human individuals), where in case of the Stochastic Logic $p(\phi \land \psi)$ can be interpreted as a product $p(\phi)p(\psi)$ with a suitable independence assumption that can be realized "externally" by "choosing a number of different individuals at random to answer each question involved in evaluating a compound". Gaines considered only propositional logic, but in the more general first-order setting of PALO we also need to define the semantics of quantifiers, which helps us to realize his idea in a quite natural fashion. Another difference to Gaines' work is that as a soft logic PALO does not satisfy the strict absorption (and hence idempotence) axioms of SUL and the additional distributivity and excluded middle axiom of Stochastic Logic (which is still a classical logic). It does so however with increasing precision when approaching the classical limit case where interpretations are constrained to $\set{0,1}$.

It is noteworthy that our approach of combining selected operators from Hajek's Product logic, \L{}ukasiewicz logic, and G\"odel logic in a non-standard fashion needs to be differentiated from work in the area of fuzzy logics, which is not aiming at a probabilistic interpretation but an orthogonal notion of truthiness (see also \cite{Gaines78} for his population-based interpretation of Fuzzy Logic). For example, \cite{Esteva1999} investigates a propositional fuzzy logic that contains Product logic, \L{}ukasiewicz logic, and G\"odel logic as sublogics and the focus is on identifying a suitable axiomatization and a class of models so that soundness and completeness can be established. In contrast, our approach with PALO is purely semantic and motivated by computational feasibility. We do not attempt to establish an axiomatic system for symbolic inference in soft logic, but rather maintain a connection to classical logic for which symbolic methods and technologies are well developed.

\textit{Incidence Calculus} \cite{Bundy85} is another approach to overcome the fact that a probabilistic interpretation of formulas is not truth-functional by using a less abstract semantics that interprets each formula as the set of assignments for which it holds so that conjunction becomes a simple intersection. Although this is an elegant solution, with our mean probability semantics that includes lower and upper bounds, it turns out that the bounds are sufficiently tight so that replacing our approximate by a strict probabilistic interpretation is unnecessary for the data-rich applications we are targeting. Two other practical difficulties with an exact probabilistic semantics are that dependencies between subformulas referring to external data are often unknown and even if all known dependencies would be taken into account it would lead to an unacceptably high computational complexity in the context of model generation and learning.

Soft logics have found renewed interest in the machine learning community, because of their potential to incorporate logical knowledge into the learning process. Most notably, \textit{Real Logic} \cite{ltns,rev-ltns} is the culmination of state-of-the-art efforts \cite{low-dim-embeddings,injecting-knowledge,knowledge-completion,fast-rel-learning,dist-repr,dist-semantics} to develop soft logic with distributional (i.e., feature-based) semantics that can be directly compiled into neural networks, specifically a subclass called  \textit{Logic Tensor Networks} (LTNs). Real Logic and its associated LTNs comprise the first framework of this kind that supports full first-order logic with functions (as opposed to relations only) without being subject to the closed-world assumption. A key innovative idea of \textit{Neural Tensor Networks} (NTNs) \cite{knowledge-completion}, that is incorporated into LTNs and the Real Logic semantics, is the use of a family of efficiently learnable continuous predicates (and functions in LTNs) represented as neural networks. This is the key to leverage deep learning technologies for model construction and enable a seamless integration. Real Logic considers truth values as degrees of satisfaction\footnote{More precisely, Real Logic and LTNs are parameterized by a t-norm, which is flexible enough to represent a whole family of soft logics, including (variations of) Product logic, \L{}ukasiewicz logic, and G\"odel logic.} and does not come with a probabilistic semantics. However, PALO can be regarded as a probabilistic variation of Real Logic, and we generalized its implementation in terms of LTNs to serve as a suitable basis for the PALO prototype.

A large body of research in the intersection of machine learning and logic has been conducted in the context of \textit{Markov Logic Networks} \cite{mln} which are equipped with a semantics in terms of \textit{Markov Random Fields}. They are also the basis for SRI's \textit{Probabilistic Consistency Engine} (PCE) \cite{pal,pce-doc,pce}. Model sampling and counting are used to assess the degree of satisfaction of theories with weighted rules, where weights can be learned from data \cite{mln-learning}. Limitations of these approaches include the inherent closed-world assumption (which is too limiting for general knowledge representation), lack of expressiveness and extensibility (e.g., no explicit probabilities, no functions, no equations), computationally expensive MCMC model sampling and weight learning (the model space is discrete leading to problems with state space explosion), and incompatibility with deep learning technologies (e.g., difficult to integrate with black-box deep learning components and not clear how to take advantage of massively parallel computing technologies such as GPUs). 

A fairly recent improvement
is \textit{Probabilistic Soft Logic} with \textit{Hinge-Loss Markov Random Fields} as models \cite{Bach17}. The study \cite{Lee16} shows that the relation between Markov Logic and Probabilistic Soft Logic is analogous to the relation between
Classical (Boolean) Logic and Fuzzy Logic. As in Markov Logic Networks, given a logical theory, the space of models is defined by a parameterized family of Gibbs distributions using potentials derived from (weighted) logical rules. Thanks to the use of a soft logic (specifically \L{}ukasiewic logic), the maximum a posteriori distribution can be efficiently computed by solving a convex optimization problem. While this approach is mathematically and computationally appealing, it is still based on a closed-world assumption, the use of Gibbs distributions is a significant limitation, and the relation between the distribution and the logical axioms is not as direct as desirable for a probabilistic logic. The potentials are based on \L{}ukasiewicz logic which does not have a standard probabilistic interpretation, and the precise impact of weights that are associated with the axioms is difficult to predict. PALO uses a more direct and intuitive mean probability semantics for logical theories. The tradeoff is that it does not constrain the model distribution to a known well-behaved family, and hence we are paying the price of dealing with non-convex albeit continuous optimization problems. Fortunately, we are in a position to take advantage of a broad range of algorithms for Stochastic Gradient Descent (SGD) optimization (such as \cite{Kingma14}) and Markov Chain Monte Carlo (MCMC) sampling methods leveraging SGD (such as Bayesian learning via Stochastic Gradient Langevin Dynamics \cite{Welling11,Li16}) that are available and still being further advanced due to the rapidly growing demands of deep learning architectures (see, e.g., \cite{Mandt17}, which shows how SGD can be viewed as approximate Bayesian inference).

\section{Syntax and Semantics of PALO}\label{sect-palo}

PALO is still work in progress. For clarity, we focus on the current design for the core language and semantics of PALO in this section. The full semantics of PALO combines a mean probability interpretation of formulas with upper and lower bounds and will be introduced in several stages. In addition, two types of classical semantics with different degrees of abstraction will defined to highlight, quantify, and take advantage of the connection to classical logic. A number of practically important extensions will be discussed in Section \ref{sect-ext}, and for most parts our definitions should be sufficiently general to accommodate these extensions with minor modifications.

\subsection{Syntax of the Core Logic}

A \textit{type signature} $\Sigma$ is defined by
a finite set of \textit{data types} denoted by $DType(\Sigma)$ and the distinct \textit{propositional type} $\prop \notin DType(\Sigma)$ (the type of formulas).

The set of types over $\Sigma$, denoted by $Type(\Sigma)$, is inductively defined:
$Type(\Sigma)$ contains all data types in $DType(\Sigma)$,
all \textit{(Cartesian) product types} of the form $T_1 \ldots T_n$ for $T_1, \ldots, T_n \in DType(\Sigma)$,
all \textit{function types} of the form  $T_1 \ldots T_n \rightarrow T$, and 
all \textit{predicate types} of the form $T_1 \ldots T_n \rightarrow \prop$,
where $T_1 \ldots T_n \in DType(\Sigma)$ and $T \in DType(\Sigma)$. The subsets of product types,
function types, and predicate types are denoted by $CType(\Sigma)$, $FType(\Sigma)$, and $PType(\Sigma)$, respectively.
Note that product types are defined over data types and not nested (not an essential limitation) and include all data types as a subset, i.e., data types are identified with single component product types.

Given a type signature $\Sigma$ and a \textit{dimensionality specification} $D : DType(\Sigma) \rightarrow \nat^{+}$ we inductively define the interpretation of types $\sem{\bullet}_D$:
(1) $\sem{\prop}_D = [0,1]$ (interval of $\real$ denoting truth values);
(2) $\sem{T}_D = \real^{D(T)}$ for all $T \in DType(\Sigma)$;
(3) $\sem{T_1 \ldots T_n} = \sem{T_1}_D \times \cdots \times \sem{T_n}_D$;
(3) $\sem{T_1 \ldots T_n \rightarrow T}_D = \sem{T_1}_D \times \cdots \times \sem{T_n}_D \rightarrow_c \sem{T}_D$ where
$\rightarrow_c$ denotes the space of continuous functions. It should be noted that there are no types with discrete interpretations, such as the type of natural numbers, in the current version of PALO, although extensions of PALO with such more traditional types are clearly conceivable. In the current version, such discrete types need to be embedded into continuous domains.

To simplify notation, we extend the notion of a type signature $\Sigma$ so that a dimensionality specification $D : DType(\Sigma) \rightarrow \nat^{+}$ is part of $\Sigma$ for all of the following, and we simply use $D$ to refer to it
in the context of $\Sigma$. We further extend the type signature $\Sigma$ by a \textit{complexity specification} $K : PType(\Sigma) \rightarrow \nat$, that will later be used to limit the complexity of the interpretation for symbols of predicate types so that they can be efficiently learned from data.

A \textit{signature} $\Sigma'$ extends a type signature $\Sigma$ by the following  components so that all its components are pairwise disjoint:
(1) a countable set $Const(T)$ of \textit{constant symbols} for each data type $T \in DType(\Sigma)$,
(2) a countable set $Fun(T)$ of \textit{function symbols} for each function type $T \in FType(\Sigma)$,
(3) a countable set $Prop$ of \textit{propositional constant symbols}, and
(4) a countable set $Pred(T)$ of \textit{predicate symbols} for each predicate type $T \in PType(\Sigma)$. The pairwise
disjointness requirement ensures that we do not have overloading, which simplifies the presentation of the semantics, but could be relaxed in future practical versions of PALO. To simplify notation,
the dimensionality and complexity specifications are lifted to symbols through their types.
 
A \textit{signature $\Sigma$ with sorts} $Sort$ extends a signature with a countable set of sort symbols $Sort(T)$ for each product type $T \in CType(\Sigma)$ such that all components remain pairwise disjoint. In the present version of PALO, the main role of sorts is to refer to (external) data sets inside the logic (see notion of sort binding below).\footnote{The use of two semantic levels, types and sorts, is quite similar to the use of kinds and sorts in membership equational logic \cite{meq-logic} which underlies SRI's Maude system \cite{maude}.}
A \textit{signature $\Sigma$ with variables} $Var$ extends a signature with a countable set of variable symbols $Var(T)$ for each $T \in CType(\Sigma)$ such that all its components remain pairwise disjoint. We write $\Sigma, x:T$ to denote an extended signature with a fresh $x \in Var(T)$ added to $Var$. Note that each sort, constant, function,
predicate, and variable symbol has a unique type in this setting. Note also that while constants are associated with
data types only, variables can be more generally associated with Cartesian product types, which is important for the expressiveness of quantifiers (introduced below).

Given a signature $\Sigma$ with sorts $Sort$ and variables $Var$ we define the set of \textit{terms} over $\Sigma$ and their \textit{types} inductively as follows:
(1) a variable $x \in Var(T)$ is a term of data type $T$;
(2) a constant $c \in Const(T)$ is a term of data type $T$;
(3) a \textit{tuple} $t_1 \ldots t_n$ is a term of product type $T_1 \ldots T_n$ if $t_i$ is a term of data type $T_i$;
(3) a \textit{function application} $f(t)$ is a term of type $T'$ for $f \in Fun(T \rightarrow T')$ if $t$ is a term of product type $T$;
(4) a propositional constant $\phi \in Prop$ is a term of type $\prop$;
(5) the propositions $\lfalse$ (\textit{false}) and $\ltrue$ (\textit{true}) are terms of type $\prop$;
(6) a \textit{predicate application} $p(t)$ is a term of type $\prop$ for $p \in Pred(T \rightarrow \prop)$  if $t$ is a term of product type $T$;
(7) logical \textit{conjunction} $\phi \land \psi$, \textit{disjunction} $\phi \lor \psi$, \textit{negation} $\neg \phi$,  \textit{implication} $\phi \Rightarrow \psi$, and \textit{equivalence} $\phi \Leftrightarrow \psi$ are terms of type $\prop$ if $\phi$ and $\psi$ are terms of type $\prop$;
(8) \textit{quantifications} $\all{x:s}{\phi}$ (\textit{universal quantification}), $\ex{x:s}{\phi}$ (\textit{existential quantification}), and $\mn{x:s}{\phi}$ (\textit{mean quantification}) are terms of type $\prop$ if $s$ is a sort of product type $S$ and $\phi$ is a term of type $\prop$ over $\Sigma, x : S$; and
(9) \textit{tensor abstraction} $\abs{x:s}{t}$ is a term of type $S \rightarrow T$ if $s$ is a sort of product type $S$ and t is a term of type $T$ over $\Sigma, x : S$. 
Terms of type $\prop$ are also called \textit{propositions} or \textit{formulas}. All other terms are also called \textit{proper terms}. The set of terms over $\Sigma$ of type $T$ is denoted as $Term(\Sigma,T)$.
In the quantifiers defined by (6) the variable $x$ is bound in $\phi$. We use standard definitions of bound/free variables and closed terms/formulas and we identify terms that are equivalent modulo renaming of bound variables. Note that in the current version of PALO each term has a unique type, which is not essential but mainly done to simplify the semantics.

\subsection{Approximate Probability Semantics}

Given a signature $\Sigma$, 
we define a $\Sigma$-algebra $A$ to consist of the following components:
(1) $A(T) \subseteq \sem{T}_D$ for $T \in DType(\Sigma)$;
(2) $A(c) \in A(T)$ for $c \in Const(T)$ for $T \in DType(\Sigma)$;
(3) $A(f) \in \sem{T_1 \ldots T_n \rightarrow T}_D$ for $f \in Fun(T_1 \ldots T_n \rightarrow T)$ and $A(f)(a_1,\ldots,a_n) \in A(T)$ for $a_i \in A(T_i)$; 
(4) $A(\phi) \in A(\prop)$ for $\phi \in Prop$; and
(5) $A(p) \in \sem{T_1 \ldots T_n \rightarrow \prop}_D$ for $p \in Pred(T_1 \ldots T_n \rightarrow \prop)$.
Note that our current description focuses on the core logic. If the signature is extended with built-in symbols, their interpretations should be added here.

Given a signature $\Sigma$ with sorts $Sort$ and a $\Sigma$-algebra $A$, a \textit{sort binding} for $Sort$ is a family of functions $\delta : Sort(T_1 \ldots T_n) \rightarrow 2^{\sem{T_1 \ldots T_n}_D}$ (implicitly) indexed by product types of the form $T_1 \ldots T_n$, such that $\delta(s) \subseteq A(T_1) \times \ldots \times A(T_n)$ and $\delta(s)$ is finite for $s \in Sort(T_1 \ldots T_n)$. Note that to establish a semantic connection to real world data sets we use an interpretation of sorts as finite, albeit typically large sets. Given a signature $\Sigma$ with variables $Var$ and a $\Sigma$-algebra $A$, a \textit{variable binding} for $Var$ is a family of functions $\beta : Var(T) \rightarrow \sem{T}_D$ (implicitly) indexed by product types $T$ such that $\beta(x) \in A(T)$ for $x \in Var(T)$. Given a binding $\beta$, a variable $x \in Var(T)$, and $d \in \sem{T}_D$, the \textit{updated} binding $\beta[x := d]$ is defined by $\beta[x := d](x) = d$ and $\beta[x := d](y) = \beta(y)$ if $x \not= y$. 

Let $\Sigma$ be a signature with sorts $Sort$ and variables $Var$, and $\theta$ a family of model parameters for $\Sigma$ that will be made precise incrementally.  For the presentation of the semantics, it will be useful to fix some variable naming conventions. We use $s$ to range over $Sort(T)$, $x$ to range over $Var(T)$, $c$ to range over $Const(T)$, $f$ to range over $Fun(T_1 \ldots T_n \rightarrow T)$, $\gamma$ to range over $Prop$, and $p$ to range over  $Pred(T_1 \ldots T_n \rightarrow \prop)$, for all suitable types $T$ and $T_1,\ldots,T_n$.  We will also use $t$ and $t_1,\ldots,t_n$ to range over proper terms over $\Sigma$ and $\phi$, $\psi$ to range over propositions over $\Sigma$.

A $\Sigma$-algebra $A$ is a $\Sigma,\theta$-algebra if constant, function, and predicate symbols are interpreted as follows:
$$A(c) = a_c$$
$$A(f)(v) = V_f v + b_f$$
$$A(\gamma) = a_\gamma$$
$$A(p)(v) = \sigma(U_p^\top \tanh (v^\top W_p^{[1:K(p)]} v + V_p v + b_p))$$
where $v$ ranges over all vectors consistent with the corresponding function or predicate type, $\sigma$ is the Sigmoid function, and $\tanh$ is applied elementwise, thereby lifted to vectors. We identify $v$ with its flattened representation, that is if $f$ is of type $T_1 \ldots T_n \rightarrow T$ the length of $v$ is $L(f) = D(T_1) + \cdots + D(T_n)$ and analogously for predicates $p$. We  use $\bullet^\top$ to denote transposition and try to stick close to the notation used in \cite{ltns,rev-ltns,knowledge-completion}.
Furthermore, $a_c$, $a_\gamma$, $V_f$, $b_f$ and $U_p^\top$, $W_p^{[1:K(p)]}$, $V_p$, $b_p$ are constant, function, and predicate \textit{model parameters} given by $\theta$. Recall that $K$ is the complexity specification that is part of a type signature. Assuming that $c$ is of type $T$, $a_c$ is a vector of length $D(T)$.
Assuming $f$ is of type $T_1,\ldots,T_n \rightarrow T$, $b_f$ is a vector of dimension $D(T)$, and $V_f$ is a matrix
of dimension $D(T) \times L(f)$. The parameter $a_\gamma$ is a scalar in $[0,1]$. Analogously to the functional case and assuming $p$ is of type $T_1,\ldots,T_n \rightarrow \prop$, the parameter $b_p$ is a vector of dimension $K(p)$, and the parameter $V_p$ is a matrix of dimension $K(p) \times L(p)$. The parameter $W_p^{[1:K(p)]}$ denotes a $[1,\ldots,K(p)]$-indexed family of matrices of dimension $L(p) \times L(p)$, and the expression $v^\top W_p^{[1:K(p)]} v$ in the above definition denotes a lifted version of the bilinear tensor product resulting in a $K(p)$-dimensional vector defined by $v^\top W_p^i v$ for $i \in 1,\ldots,K(p)$. Finally, the parameter $U_p$ is a vector of dimension $K(p)$. 
Such a $\Sigma,\theta$-algebra is uniquely defined by $\Sigma$ and $\theta$ and will be denoted by $A(\Sigma,\theta)$ or
simply $A(\theta)$.

The above parametric class of efficiently learnable, continuous, and differentiable predicates (with their natural representation as neural networks) is the same as in LTNs \cite{ltns,rev-ltns} and a direct generalization of the representation of binary predicates in NTNs  \cite{knowledge-completion}. The power of this representation is that it can capture complex non-linear interactions between the predicate arguments. One may think of $K$ as a hyperparameter limiting the number of such interactions and hence together with the dimensionality $D$ the complexity of the class of learnable interpretations. However, there is no reason to claim that this family is sufficiently rich for all applications. Similarly, while linear functions are an important subclass for many applications, it will be too restrictive for others and does not match the power of predicates. Other families of learnable interpretations are clearly conceivable (see Section \ref{sect-ext}), and we should rather think of the above choice motivated by LTNs as a particular instantiation of PALO.

For the following, we assume a signature $\Sigma$ with sorts $Sort$ and variables $Var$, a $\Sigma$-algebra $A$, and bindings $\delta$ for $Sort$ and $\beta$ for $Var$.

With that we inductively define the \textit{approximate probability interpretation} $\sem{\bullet}^a_{A,\delta,\beta} : Term(\Sigma,T) \rightarrow \sem{T}$ of terms (including proper terms and propositions) as follows:

$$\sem{x}^a_{A,\delta,\beta} = \beta(x)$$
 $$\sem{t_1,\ldots,t_m}^a_{A,\delta,\beta} = (\sem{t_1}^a_{A,\delta,\beta},\ldots,\sem{t_m}^a_{A,\delta,\beta}) \textrm{ for } m \geq 2 $$
 
$$\sem{c}^a_{A,\delta,\beta} = A(c)$$
$$\sem{f(t)}^a_{A,\delta,\beta} = A(f)(\sem{t}^a_{A,\delta,\beta}) $$
$$\sem{\gamma}^a_{A,\delta,\beta} = A(\gamma)$$
$$\sem{P(t)}^a_{A,\delta,\beta} = A(P)(\sem{t}^a_{A,\delta,\beta}) $$

$$\sem{\lfalse}^a_{A,\delta,\beta} = 0$$
$$\sem{\ltrue}^a_{A,\delta,\beta} = 1$$
$$\sem{\neg \phi}^a_{A,\delta,\beta} = 1 - \sem{\phi}^a_{A,\delta,\beta}$$ 
$$\sem{\phi \land \psi}^a_{A,\delta,\beta} = \sem{\phi}^a_{A,\delta,\beta} \cdot \sem{\psi}^a_{A,\delta,\beta}$$
$$\sem{\phi \lor \psi}^a_{A,\delta,\beta} =  \sem{\neg(\neg \phi \land \neg \psi)}^a_{A,\delta,\beta}  = \sem{\phi}^a_{A,\delta,\beta} + \sem{\psi}^a_{A,\delta,\beta} - \sem{\phi}^a_{A,\delta,\beta} \cdot \sem{\psi}^a_{A,\delta,\beta}$$
$$\sem{\phi \Rightarrow \psi}^a_{A,\delta,\beta} = \sem{\neg \phi \lor \psi}^a_{A,\delta,\beta} = \sem{\phi}^a_{A,\delta,\beta} \cdot \sem{\psi}^a_{A,\delta,\beta} - \sem{\phi}^a_{A,\delta,\beta} + 1$$
 $$\sem{\phi \Leftrightarrow \psi}^a_{A,\delta,\beta} = 
 \sem{\phi \land \psi}_{A,\delta,\beta}^a + \sem{\neg\phi \land \neg\psi}_{A,\delta,\beta}^a $$
$$\hspace*{4.5cm} =  1- \sem{\phi}^a_{A,\delta,\beta} - \sem{\psi}^a_{A,\delta,\beta} + 2 \cdot  \sem{\phi}^a_{A,\delta,\beta} \cdot \sem{\psi}^a_{A,\delta,\beta}$$

$$\sem{\ex{x : s}{\phi}}^a_{A,\delta,\beta} = \max_{d \in \delta(s)} \sem{\phi}^a_{A,\delta,\beta[x:=d]}$$

$$\sem{\all{x : s}{\phi}}^a_{A,\delta,\beta} = (\prod_{d \in \delta(s)} \sem{\phi}^a_{A,\delta,\beta[x:=d]})^ \frac{1}{\card{\delta(s)}}$$

$$\sem{\mn{x : s}{\phi}}^a_{A,\delta,\beta} = \frac{1}{\card{\delta(s)}}\sum_{d \in \delta(s)} \sem{\phi}^a_{A,\delta,\beta[x:=d]}$$

$$\sem{\abs{x:s}{t}}^a_{A,\delta,\beta} = d \in \delta(S) \mapsto \sem{t}^a_{A,\delta,\beta[x:=d]}$$

Note that while conjunction is defined as in Hajek's Product logic \cite{Hajek96}, the involutive negation is defined as in \L{}ukasiewicz logic and implication has the standard classical definition\footnote{also called S-implication which is different from Product logic's R-implication}. For equivalence, we use the probabilistically more accurate definition consistent with \cite{Gaines78} rather than defining it as a derived operator. The meaning of conjunction is an exact probability if its subformulas are statistically independent, and should otherwise be seen as a best guess or approximate probability in absence of information about their dependence.\footnote{A more accurate explanation is that there is an implicit assumption that the subformulas are semantically sufficiently diverse so that a potential dependence is negligible relative to the diversity caused by different instantiations.} It's (average) precision will be made more precise in the full semantics which contains lower and upper bounds.

Disjunction is defined using De Morgan's laws, establishing full symmetry between conjunction and disjunction. Associativity holds for both. However, the substructural (specifically linear) nature of this soft logic manifests itself by the fact that idempotence, and consequently absorption and distributivity are not valid.\footnote{It is noteworthy that already in \cite{Gaines78} it is pointed out that a case can be made for weaker systems of his Standard Uncertainty Logic (SUL) without idempotence, which is the only reason for the lack of these properties in PALO.} The law of the excluded middle and the law of contradiction are equivalent, because $\sem{\phi \lor \neg \phi} + \sem{\phi \land \neg \phi} = 1$ holds just as in \cite{Gaines78}, but neither is valid in PALO. However, idempotence and hence all of  these properties are valid in the classical limit case where formulas are interpreted in $\set{0,1}$, which will be the case in the alternative classical semantics for PALO. A key property identified in \cite{Gaines78} that also holds
in PALO in spite of the lack of idempotence is 
$$\sem{\psi} = \sem{\phi \lor \psi} -1 + \sem{\phi \Rightarrow \psi} \geq \sem{\phi} - (1 - \sem{\phi \Rightarrow \psi})$$
which allows a limited form of modus ponens in the sense that it enforces a lower bound for $\sem{\psi}$ given $\sem{\phi}$
and $\sem{\phi \Rightarrow \psi}$, but as we will see this is only one form of inference that can take place in PALO
which unlike most deductive systems does not favor any particular direction of execution.

The existential quantifier, on the other hand, is defined in terms of the maximum,
which is consistent with Skolemnization (the most typical and intuitive classical interpretation) 
if the set of functions was sufficiently rich (which is not the case in our basic semantics, and another motivation for
considering larger classes of functions). The universal quantifier is interpreted as a particular geometric mean, which is consistent with a (normalized) product interpretation of the quantifier viewed as a large conjunction over a batch of data. Here we need the assumption of (approximate) independence for the elements within each batch, which in practice limits the maximum batch size (see next subsection). Clearly, universal and existential quantifiers are not duals of each other, but for each quantifier we could formally define its dual counterpart by $\dall{x:s}{\phi} = \neg \all{x:s}{\neg \phi}$ and $\dex{x:s}{\phi} = \neg \ex{x:s}{\neg \phi}$ giving rise to four different quantifiers, thereby reestablishing a formal symmetry.\footnote{More experience is needed to determine if the dual quantifiers are of any practical use.} The mean quantifier is quantitatively between universal and existential
quantifiers and most directly captures the arithmetic mean probability of a formula.

Also note that the tensor abstraction could be used to define all three quantifiers, but for clarity we have used an explicit definition here. Hence, in the core logic we are presenting, the tensor abstraction cannot appear as an argument, but with suitable built-in (higher order) function symbols that would be possible in an extension of the core logic.

\subsection{Mean Probability Semantics}

Given a signature $\Sigma$ with sorts $Sort$, a $\Sigma$-algebra $A$, and a sort binding $\delta$ for $Sort$, a \textit{batch cover} is a  function $\Delta$ with $\Delta(\delta,s) \subseteq 2^{\delta(s)}$ and $\delta(s) = \bigcup \Delta(\delta,s)$ for each sort $s \in Sort(T)$ and $B \subseteq B'$ implies $B = B'$ for all $B,B' \in \Delta(\delta,s)$. 
A possible choice is the set of all subsets of $\delta(s)$ of a fixed size. 
Lifting this concept from sorts to sort bindings, we define the \textit{batch cover} $\Delta(\delta)$ as a set of all sort bindings $\delta'$ such that $\delta'(s) \in \Delta(\delta,s)$ for each $s \in Sort$. 
Using $\Delta$ to generically denote any set of sort bindings for $Sort$ under $A$ (typically it will be $\Delta(\delta)$), 
this gives rise to the \textit{mean probability interpretation}
of closed formulas $\phi$:

$$\sem{\phi}^a_{A,\Delta} = \frac{1}{\card{\Delta}}\sum_{\delta' \in \Delta} \sem{\phi}^a_{A,\delta'}$$

The mean probability semantics can seen as an approximation of the population semantics of the Stochastic Logic in \cite{Gaines78} for formulas that exhibit enough statistical diversity. Two important distinctions are however that we have to deal with quantifiers and we use a soft logic rather than a classical one.

While we average over all possible combinations $\Delta(\delta)$ of batches  for all sorts in the above definition, this is usually not feasible in practice, and hence a natural place where an implementation would approximate the sum by using a random subset of $\Delta(\delta)$, where each sort is interpreted by batches of a fixed sample size.  The determination of a suitable batch size for the application is left as a topic for  further investigation. A rough although incomplete guide can be the concept of effective sample size from statistics.

For the following, we assume a signature $\Sigma$ with sorts, a $\Sigma$-algebra $A$, and a set $\Delta$ of sort bindings under $A$ (constructed from a batch cover as above). For a closed formula $\phi$ we say that $A$ satisfies $\phi$ or $A$ is a \textit{model} of $\phi$ with \textit{lower mean probability} $l$ and \textit{upper mean probability} $u$ (denoted by $A \models^u_l \phi$) iff $\sem{\phi}^a_{A,\Delta} \in [l,u]$. In the following we lift this notion to theories.

\subsection{Probabilistic Theories and their Semantics}

A \textit{probabilistic theory} $\Gamma$ is a set of triples $(l,\phi,u)$, where $\phi$ is a closed formula over $\Sigma$, and $l,u \in [0,1]$ with $l \leq u$ define a probability interval $[l,u]$ for the truth value of $\phi$. We say $\Sigma$-algebra $A$ satisfies $\Gamma$ or  $A$ is a model of $\Gamma$ (denoted by $A \models \Gamma$) iff $A \models^u_l \phi$ for all $(l,\phi,u) \in \Gamma$. We write $\phi \in \Gamma$ if $(l,\phi,u) \in \Gamma$ for some $l$,$u$, and call $\phi$ an \textit{axiom} of $\Gamma$. 

We extend the interpretation $\sem{\bullet}^a_{A,\Delta}$ to probabilistic theories as follows:
$$\sem{\Gamma}^a_{A,\Delta} = \prod_{(l,\phi,u) \in \Gamma} \sem{(l,\phi,u)}^a_{A,\Delta}$$
where $\sem{(l,\phi,u)}^a_{A,\Delta} = \sem{\phi}^a_{A,\Delta}$ if $l \leq \sem{\phi}^a_{A,\Delta} \leq u$ else $\sem{(l,\phi,u)}^a_{A,\Delta} = 0$.

\vspace*{1ex}

While in the previous definition the $\Sigma$-algebra $A$ was unrestricted, we now return to the class of algebras $A(\theta)$ generated by model parameters $\theta \in Param(\Sigma)$, where $Param(\Sigma)$
denotes the set of all model parameter instantiations. Similarly, we assume that $\Delta$ is of the form $\Delta(\delta)$, meaning that it is generated by a batch cover $\Delta$ and a sort binding $\delta$ representing the data.

A probabilistic theory $\Gamma$ is \textit{satisfiable} if $\sem{\Gamma}^a_{A(\theta),\Delta(\delta)} > 0$ for some $\theta \in Param(\Sigma)$. For a satisfiable probabilistic theory $\Gamma$, the \textit{maximum likelihood model} is defined by the following optimization problem: 
$$\theta^\ast = \argmax_{\theta \in Param(\Sigma)} \sem{\Gamma}^a_{A(\theta),\Delta(\delta)} $$

The simple product formulation above, however, is too inflexible to account for the complexity of logical theories used in practical applications that by their very nature often involve complex dependencies that cannot be eliminated or sufficiently reduced by exploiting randomization or diversity. Hence, this formulation only serves as a stepping stone to a more general approach that introduces flexibility at the level of theories, which in a sense is dual to and complements the flexibility at the level of models that we already take for granted.

A \textit{flexible probabilistic theory} $\Gamma$ is a family of probabilistic theories parameterized by $Param(\Sigma)$ such that $\Gamma(\theta)$ 
is a set of triples of the form $(l,\phi^r,u)$ where $r = w_\theta(\phi)$ is a parameter in $\theta$ specific to $\phi$, also called a (flexible) \textit{weight} for $\phi$. 
Together, all parameters $w_\theta(\phi)$ are called \textit{theory parameters} to distinguish them from the \textit{model parameters}
defined previously. From now on all instantiations for both types of parameters will be included in $Param(\Sigma)$.

We now extend the interpretation $\sem{\bullet}^a_{A,\Delta}$ to flexible probabilistic theories as follows:
$$\sem{\Gamma(\theta)}^a_{A,\Delta} = \prod_{(\phi^r,l,u) \in \Gamma} \sem{(\phi^r,l,u)}^a_{A,\Delta}$$
where $\sem{(\phi^r,l,u)}^a_{A,\Delta} = (\sem{\phi}^a_{A,\Delta})^r$ if $l \leq \sem{\phi}^a_{A,\Delta} \leq u$ else $\sem{(\phi^r,l,u)}^a_{A,\Delta} = 0$. Note that the constraints are on $\phi$ rather than $\phi^r$, which
means that the weights are irrelevant for the probabilistic interpretation and constraints for individual axioms, 
but rather control how axioms are composed. With flexible weights PALO tries to (partially) compensate
for complex dependencies between axioms. We will see later that such dependencies will be
unavoidable and in fact essential in a context where classical logic is our reference for approximation.\footnote{This is in contrast to attempts to base theories on minimal axiomatizations which is beneficial for inductive arguments in the proof theory of logics.}

A flexible probabilistic theory $\Gamma$ is \textit{satisfiable} if $\sem{\Gamma(\theta)}^a_{A(\theta),\Delta(\delta)} > 0$ for some $\theta \in Param(\Sigma)$.
For a satisfiable flexible probabilistic theory the \textit{maximum likelihood theory and model} is defined by the following optimization problem:
$$\theta^\ast = \argmax_{\theta \in Param(\Sigma)} \sem{\Gamma(\theta)}^a_{A(\theta),\Delta(\delta))} $$

In general, however, the interpretation of a theory is a non-convex function admitting many local
maxima, some may be caused by symmetries in the theory or in the interpretation of the symbols.\footnote{More precisely, there is no guarantee of consistency for the likelihood function associated with (flexible) theories.} We might also
have prior knowledge about the distribution of parameters or enforce certain types of regularization, e.g., to balance model complexity and the amount of available data. In such case, a Bayesian treatment is more appropriate, where a flexible probabilistic theory $\Gamma$ induces a conditional probability (proportional to the likelihood\footnote{We are dealing here with an unnormalized likelihood, because like complex systems in statistical physics, our interpretation of theories contains an unknown normalization constant whose computation is infeasible in practice but fortunately irrelevant for the semantics.})

$$p(\delta\vert\theta) \propto L(\theta ; \delta) = \sem{\Gamma(\theta)}^a_{A(\theta),\Delta(\delta)} $$
and the joint probability factorizes as
$$p(\theta,\delta) = p(\delta\vert\theta) p(\theta) = p(\theta\vert\delta) p(\delta)$$
if $p(\theta)$ is an assumed prior for the parameters $\theta$.

Given a flexible theory $\Gamma$ and a prior $p(\theta)$ for all parameters $\theta \in Param(\Sigma)$, we can now define three flavors of Bayesian semantics (which have to be approximated in practice). The \textit{posterior mode semantics} is the set of all triples $(\theta,\Gamma(\theta),A(\theta))$ such that $p(\theta \vert \delta)$ is a local maximum w.r.t. $\theta$. The \textit{maximum a posteriori semantics} is the set of all triples $(\theta,\Gamma(\theta),A(\theta))$ such that  $p(\theta \vert \delta)$ is a global maximum w.r.t. $\theta$. Note that we allow a set of global optima instead of only a single one to account for symmetries. Finally, the most general \textit{posterior distribution semantics} is the set of all triples $(\theta,\Gamma(\theta),A(\theta))$ equipped with a probability distribution $p((\theta,\Gamma(\theta),A(\theta))) \propto p(\theta \vert \delta) \propto  p(\delta\vert\theta) p(\theta)$. 

Depending on the intended applications, several variations are possible in this framework such as the following staged hybrid semantics. A \textit{posterior model distribution semantics} is  the set of all triples $(\theta,\Gamma(\theta),A(\theta))$ with a probability distribution $p(\theta_m \vert \theta_t,\delta)$ where $\theta_m$ are the model parameters and the theory parameters $\theta_t$ are determined by the maximum a posteriori semantics.

Any form of statistical inference related to the different flavors of posterior semantics will also be called
\textit{(approximate) computational inference} to distinguish it from \textit{(exact) symbolic inference} as
it is traditionally used in logics.

\subsection{Lower and Upper Probability Semantics}

The approximate probability semantics $\sem{\bullet}^a_{A,\delta,\beta}$ gives an exact probability only if
the subformulas of composite formulas are independent which is rarely the case in practice. Consider,
for example the extreme case of an atomic formula $\phi$ with $\sem{\phi}^a = 0.5$. The approximate semantics
yields $\sem{\phi \land \neg \phi}^a = 0.25$ even though classically the formula is equivalent to $\lfalse$. Another extreme example exploiting the lack of idempotence is $\sem{\phi \land \phi}^a = 0.25$.
A semantics that captures this propositional imprecision by interpreting formulas in terms of probability
intervals can be defined as follows.

A \textit{lower and upper probability semantics} (which we also call Frech\'e semantics) can be obtained by using
Frech\'e bounds as follows:

$$\sem{\neg \phi}^l_{A,\delta,\beta} = 1 - \sem{\phi}^u_{A,\delta,\beta}$$ 
$$\sem{\neg \phi}^u_{A,\delta,\beta} = 1 - \sem{\phi}^l_{A,\delta,\beta}$$ 

$$\sem{\phi \land \psi}^l_{A,\delta,\beta} = \max(0,\sem{\phi}^l_{A,\delta,\beta} + \sem{\psi}^l_{A,\delta,\beta} - 1)$$
$$\sem{\phi \land \psi}^u_{A,\delta,\beta} = \min(\sem{\phi}^u_{A,\delta,\beta}, \sem{\psi}^u_{A,\delta,\beta})$$

$$\sem{\phi \lor \psi}^l_{A,\delta,\beta} = \max(\sem{\phi}^l_{A,\delta,\beta}, \sem{\psi}^l_{A,\delta,\beta}) $$
$$\sem{\phi \lor \psi}^u_{A,\delta,\beta} = \min(1, \sem{\phi}^u_{A,\delta,\beta} + \sem{\psi}^u_{A,\delta,\beta}) $$

$$\sem{\phi \Rightarrow \psi}^l_{A,\delta,\beta} = \sem{\neg \phi \lor \psi}_{A,\delta,\beta}^l $$
$$\sem{\phi \Rightarrow \psi}^u_{A,\delta,\beta} = \sem{\neg \phi \lor \psi}_{A,\delta,\beta}^u $$

$$\sem{\phi \Leftrightarrow \psi}^u_{A,\delta,\beta} = \sem{\phi \land \psi}_{A,\delta,\beta}^u + \sem{\neg\phi \land \neg\psi}_{A,\delta,\beta}^u $$
$$\sem{\phi \Leftrightarrow \psi}^l_{A,\delta,\beta} = \sem{\phi \land \psi}_{A,\delta,\beta}^l + \sem{\neg\phi \land \neg\psi}_{A,\delta,\beta}^l $$

All other equations from the definition of $\sem{\bullet}^a_{A,\delta,\beta}$ are duplicated for 
$\sem{\bullet}^l_{A,\delta,\beta}$ and $\sem{\bullet}^u_{A,\delta,\beta}$  without changes. The same
holds for the mean probability interpretation $\sem{\phi}^a_{A,\Delta}$, which is duplicated
for $\sem{\phi}^l_{A,\Delta}$ and $\sem{\phi}^u_{A,\Delta}$. The independence 
assumptions for batches are still required under this semantics.

\vspace{0.5ex}

Note that while $\sem{\bullet}^a_{A,\delta,\beta}$ defines conjunction as in Hajek's Product logic and disjunction using involutive negation, $\sem{\phi \land \psi}^l_{A,\delta,\beta}$ and $\sem{\phi \lor \psi}^u_{A,\delta,\beta}$ are defined as (strong) conjunction and disjunction in \L{}ukasiewicz logic, while $\sem{\phi \land \psi}^u_{A,\delta,\beta}$ and $\sem{\phi \lor \psi}^l_{A,\delta,\beta}$ correspond to G\"odel logic (also called weak conjunction and disjunction in \L{}ukasiewicz logic). 

Since lower and upper probabilities are essential to understand the precision of the approximate semantics,  
it is natural to extend $\sem{\phi}^a_{A,\delta,\beta}$ and $\sem{\phi}^a_{A,\Delta}$ to the combined semantics based on triples:

$$\sem{\phi}_{A,\delta,\beta} = (\sem{\phi}^l_{A,\delta,\beta},\sem{\phi}^a_{A,\delta,\beta} ,\sem{\phi}^u_{A,\delta,\beta} )$$
$$\sem{\phi}_{A,\Delta} = (\sem{\phi}^l_{A,\Delta},\sem{\phi}^a_{A,\Delta} ,\sem{\phi}^u_{A,\Delta} )$$

Applying this extended semantics to the extreme example $\phi \land \neg \phi$ we obtain
$\sem{\phi \land \neg \phi} = (0,0.25,1)$. While this does not give us a better estimate, it alerts us about
the high degree of imprecision in the approximate probability. Typically, formulas appear in the context of quantifiers,
resulting in reasonably narrow bounds for the approximate mean probability. However, if the interval is still too large,
it can be a sign of possible inherent dependencies between subformulas and a reformulation of the formula would be a natural response, for example using classical reasoning, which would yield $\lfalse$ in our example.

\subsection{Abstract Classical Semantics for Reasoning}

An \textit{abstract classical semantics} (for the fragment without the mean quantifier) can be obtained by 
restricting $\sem{\prop}$ to the Boolean set $\set{0,1}$ and using a trivial
batch cover $\Delta(\delta,s) = \set{\delta(s)}$.  
The latter means that the semantics of quantifiers is exact rather than
approximate, which suggests that computationally this semantics will not be useful in most practical cases. However,
the use of a simple abstract semantics, justifies the use of symbolic deduction. Although formally a special case, we use $\sem{\phi}^c_{A,\delta,\beta}$ and $\sem{\phi}^c_{A,\Delta}$ instead of $\sem{\phi}^a_{A,\delta,\beta}$ and $\sem{\phi}^a_{A,\Delta}$ to make clear that we are using the classical semantics.

Let $A$ be an arbitrary $\Sigma$-algebra and $\phi$ a closed formula not containing the mean quantifier.
We say that $A$ \textit{satisfies} $\phi$ or $A$ is a \textit{model} of $\phi$ (denoted by $A \models^c \phi$) iff 
$\sem{\phi}^c_{A,\Delta} = 1$. We say $A$ \textit{satisfies} $\Gamma$ or $A$ is a \textit{model} of $\Gamma$ (denoted by $A \models^c \Gamma$) iff $A \models^c \phi$ for all $\phi \in \Gamma$. We say $\phi$ is a \textit{tautology} of $\Gamma$, written $\Gamma \models^c \phi$, iff $A \models^c \phi$ for all $\Sigma$-algebras $A$ such that $A \models^c \Gamma$. 
Note that the classical semantics uses all $\Sigma$-algebras $A$, not a parameterized subset
such as $A(\theta)$. Classical tautologies can be established symbolically, e.g., as theorems generated by a 
sound and complete\footnote{It should be noted that completeness does not imply completeness for our approximate semantics, because the class of predicates and functions is unrestricted in the abstract classical semantics. Hence soundness is the more important property here.} proof system for first-order logic, but the particular method is not relevant for this discussion.

A (flexible) probabilistic theory should be regarded as an approximation of a classical theory in the
sense that starting from a core theory $\Gamma_0$ we can generate potentially infinite chains of 
theories using symbolic deduction 
$$ \Gamma_0 \subseteq \Gamma_1 \subseteq \Gamma_2 \subseteq \cdots $$
that are all classically equivalent but not necessarily equivalent under the probabilistic approximate semantics.\footnote{Our argument naturally extended to a chain of embeddings, which allow us to introduce new (auxiliary) functions and predicates.} 

Due to the inherent limitations of a soft logic to mimic exact inference, 
instead of using $\Gamma_0$ as the basis for probabilistic approximate inference the use of 
$\Gamma_i$ for some $i > 0$ can lead to substantial improvements in efficiency and precision
in approximating classical reasoning. An equivalent perspective is that in addition to the axiomatic domain
knowledge additional classical theorems can be made available to the probabilistic engine. The determination
of a suitable set of theorems is an interesting problem by itself and most likely related to a tradeoff between
computational efficiency and precision that should be further investigated. Note that the classical
semantics completely abstracts from the probabilities, which means both the approximate as well as the 
classical semantics are limited in their own ways, and it would be inappropriate to consider one superior over the other.

The complementary nature of computational inference using the approximate semantics and symbolic inference using the classical semantics is an interesting topic by itself that, although beyond the scope of this paper, leads to the idea of hybrid neural-symbolic architectures (to be briefly discussed in Section \ref{sect-ext}) that integrate both forms of reasoning in a synergistic way. The advantage of computational inference is its non-sequential, non-directed, and fuzzy nature that can benefit from today's massively parallel hardware architectures. Symbolic inference, on the other hand, can maintain logical precision over many reasoning steps and at the same time work with templates of formulas or entire classes of models, essentially leading to a logical form of parallelism by exploiting symmetries and abstractions.

\subsection{Concrete Classical Semantics for Validation}

A less abstract \textit{probabilistic classical semantics} can be obtained by crispification.\footnote{The term crispification is inspired by \cite{Lee16}, albeit in that reference it is enforced as an axiom, while we define it at the semantic level.} This can be useful if a soft-logic model has already been identified, and we would like to reinterpret the model from a classical viewpoint.

To this end, we define a \textit{crispification} operator $\crisp_\tau : [0,1] \rightarrow \set{0,1}$, where $\tau \in [0,1]$ is a fixed threshold, by $\crisp_\tau(\phi) = 1$ if $\phi \geq \tau$ and $0$ otherwise.
Now $\textrm{crisp}_\tau(\phi)$ is inserted in each equation for $\sem{\bullet}^a_{A,\delta,\beta}$ dealing with a term of propositional type. The resulting semantics is the mean probability interpretation $\sem{\phi}_{A,\Delta}$ for a $\Sigma$-algebra $A$ and batch cover $\Delta$, which we denote as $\sem{\phi}^\tau_{A,\Delta}$ or simply
$\sem{\phi}^{0.5}_{A,\Delta}$ using the default value for $\tau$.

While crispification itself leads to a loss of precision regarding the model, the use of classical logic, on the other hand, increases logical precision (by avoiding the incompleteness of the soft-logic approximation). This is a tradeoff that may give some insights into the structure of a given model and motivate new hypothesis and extensions of the domain theory. It is also a reference to validate the standard semantics against, e.g., to quantify the degree of incompleteness in the context of
specific theories and applications.

\section{The Logical Imagination Engine}\label{sect-lime}

A partial prototype of the Logical Imagination Engine (LIME) has been developed using a generalized and extended
version of Logic Tensor Networks (LTNs), which are implemented as a layer on top of TensorFlow \cite{tensorflow}. Currently it implements the posterior mode semantics (including lower and upper probabilities) using model sampling with Adam \cite{Kingma14} and also the concrete classical semantics for a given model. We expect that implementing the full posterior distribution semantics would be straightforward using SGD/Langevin MCMC sampling \cite{Li16} that is already available in TensorFlow. Our LIME prototype operates within JupyterFlow \cite{canes,Vertes18}, our notebook-based distributed workflow framework for Python/TensorFlow that transparently takes advantage of clusters of heterogeneous machines (e.g., with varying numbers of CPUs and GPUs).\footnote{Our latest version also supports virtual kubernetes clusters and takes advantage of special features in the Google cloud to efficiently share large amounts of data.} We currently use a cluster of machines to parallelize model sampling and other application-specific tasks (such as graph synthesis in our bioinformatics application discussed in the next section).

There are some minor limitations of our prototype that do not lead to severe restrictions in practice. For example, no type checking is implemented yet and each variable is associated with a unique fixed sort (i.e., we can view the variable name as the sort). Also, a temporary limitation is that axioms have to be in a particular form (essentially negation is pushed down to the level of atomic propositions) to make use of the upper and lower bounds semantics. Finally, we use a restricted family of batch covers (used in the sampling semantics for quantifiers) that are parameterized by sample size (number of random sort-bindings) and a sort-specific batch size that should be sufficient for most applications. One limitation that is quite significant is that the current prototype is inheriting the linear semantics for functions from LTNs, leading to a mismatch with the quite rich interpretation of predicates, and it does not allow for type-dependency of the complexity specification. This will be easy to rectify, but needs to be carefully evaluated in the context of an application that makes use of more complex functions such as multivariate polynomials (our current bioinformatics application is using predicates only).

With these limitations in mind, LIME implements the following functionality.\footnote{It is based on an extension of the LTN API which well-structured and easy to use.} A signature can be defined by listing symbols for constants, predicates, and functions with their type (currently only their dimension needs to be specified). A theory is defined by listing the axioms, where each axiom can be equipped with lower and upper mean probability constraints. The constraints are taken into account by a suitable extension of the TensorFlow objective function, which without constraints is simply given by the likelihood (defined by the mean probability semantics). Note that the mean probability semantics is dependent on model and theory parameters, which are translated into TensorFlow variables behind the scenes.\footnote{To represent flexible theories, flexible axioms are used, but for experimentation we also support the option of using fixed weight axioms, in which case the weight has to be specified explicitly. We do not recommend its use, however, due to the non-intuitive behavior of weights.}

Given a signature and a theory, LIME supports \textit{model synthesis} (also called learning or training) and \textit{model analysis} (also called querying, but we distinguish between \textit{model validation}\footnote{This is similar to model checking, which verifies if a given model satisfies a given property, but in PALO models are learned from a logical theory and data.} and \textit{model evaluation}). \textit{Model sampling} is the (repeated) use of model synthesis to generate a distribution of models, which is strongly biased towards maximum likelihood in the posterior mode semantics.  All functions are parameterized by a binding $\delta$ of sorts to their associated domains (e.g., data sets for training or validation), which defines the set of sample sort-bindings $\Delta(\delta)$
and forms the basis of our semantics.

Model synthesis has additional parameters such as the maximum number of training epochs (maximum number of sample sort-bindings used for training), a patience parameter for early stopping if no progress is made for the specified number of epochs (to reduce overfitting or overthinking as it might be called in this context of a logical theory), a minimum likelihood threshold to discard models with lower likelihood, and a maximum number of trials to find a model above this threshold. If successful, model synthesis results in an implicitly stored model that can be subjected to further model analysis.

Model analysis comes in two flavors. In both cases, the currently active model is an implicit parameter. Model validation (based on incomplete sampling) computes the mean probability semantics of a formula (using sampling for the quantifiers). A sort binding and a sample size (that is the number of sample sort-bindings over which the mean is taken) is passed as a parameter. Model evaluation (based on exhaustive iteration) considers all free variables of the given formula implicitly bound by the tensor abstraction $\lambda$ and hence, under a given sort binding, results in a tensor with one dimension for each free variable. This provides a natural way to extract detailed information from the model, e.g., an exhaustive enumeration of the probabilistic interpretation of a predicate for a finite set of arguments. Note that model evaluation
does not use the mean probability semantics. Instead of using $\delta$ to generate sample sort-bindings $\Delta(\delta)$, it directly uses $\delta$ to perform the evaluation, which is typically exhaustive for the domains of interest.

It should be noted that the combination with an expressive logic naturally leads to a generalization of the traditional notion of validation in machine learning (e.g., the simple notion based on a separation of training and test sets). With PALO we are concerned with two orthogonal dimensions of generalization: the generalization of a property from the training to the test set, and the generalization of an axiom (which has been used during training) to another property (which is used during testing only). LIME's model validation functionality is sufficiently general to support both notions and a combination of these. In addition, there is another dimension of validation offered by the flexible semantics and its configuration at runtime (see below).

In the following we summarize some of the generalizations/extensions of the LTN library that were necessary for the implementation of the LIME prototype. The LTN syntax was extended with a mean operator, as it needs to be clearly distinguished from universal and existential quantifiers. The parameterization has been extended to include PALO as a logic together with a definition of its non-standard semantics.  As part of the PALO semantics, we added sampling, leading to new implementations of model synthesis and analysis with the parameters mentioned above. Furthermore, the selection of the semantics has been made dynamic, so that existing models can be viewed or reinterpreted under different semantics, simply by switching the semantics at runtime. In this configurable framework, we also added the lower/upper probability semantics and the concrete classical semantics based on crispification.

As an experimental feature we have implemented an alternative approximation of the posterior mode semantics. It exploits
the new capability of switching the semantics at runtime, which is supported even during the training process. Inspired by the notion of curriculum learning \cite{curriculum-learning}, which proceeds in stages of increasing complexity of the training data, we consider our semantics as a limit case of a chain of semantics that differ in the interpretation of the existential quantifier. A staged training schedule with increasing semantic complexity can avoid the computationally fragile maximum interpretation of the existential quantifier early in the training process\footnote{In fact, training starts with an interpretation that matches $\bar{\exists}$ that has been earlier mentioned as a dual of $\forall$ and hence might shed some light on its role.} and has the potential to improve stability and efficiency of model synthesis, but more experience with applications is needed and a detailed evaluation has to remain a topic for future work.

\section{Sample Bioinformatics Application}\label{sect-bio}

In the DARPA \textit{Rapid Threat Assessment} (RTA) program we have been developing data analysis, machine-learning, and logic-based techniques to support biologists in understanding the so-called \textit{mechanism of action} (MoA) that is triggered when an (unknown) drug or toxic substance hits a human cell. From relatively short windows of time after the event in question (e.g., 48 hours) our algorithms generate graphs representing potential causality between compounds.\footnote{In spite of the use of perturbations, it should be noted that this abstract notion of causality is based on observational data with its known limitations (e.g., confounding effects), and might be better called causality modulo observational equivalence. This is in contrast to for example knock-out studies for individual genes, which however due to higher cost cannot compete with the sheer data volume and coverage typical for observational studies.} The basis are time series of typically high-dimensional data, e.g., transcriptomics (gene expression), proteomics, and metabolomics data. We have also developed algorithms for anomaly detection that highlight certain nodes in such graphs as
potentially impacted and allow the biologist to narrow down the mechanism of action. The algorithms developed use a
variety of models including Gaussian processes (on non-linear time scales) and other linear and non-linear models, ranging from principal component analysis and various types of clustering to a broad range of neural network models. For more details about the project and some initial results we refer to \cite{Vertes18}.

More recent algorithms that we developed include anomaly detection using convolutional autoencoders, autoencoder-based causality detection and network graph synthesis, predictive deep neural networks for temporal evolution and their visualization as graphs, generative adversarial networks for synthetic modeling and detection of typical vs.\ unusual behavior, and Siamese (twin) neural networks for probabilistic causality detection (validated using a dynamic gene expression model taking advantage of our original Gaussian process model). An informal presentation of our causal network synthesis algorithms and some sample results for the RTA data can be found in \cite{canes}.

One challenge that we encountered is that each type of algorithm has its own representation of biological assumptions. For example, autoencoder-based causal network synthesis makes some assumptions about the nature of biological causality, which are hardwired into the algorithm. This is not only unsatisfactory from an engineering point of view but also leads to limitations and inflexibility regarding the kind of domain knowledge that can be represented.

In the latest generation of algorithms we used PALO to represent biological domain knowledge as a logical theory.
A domain theory of causality specific to the biological domain is used as background knowledge during learning, resulting in an entire distribution of models that are probabilistically consistent with the theory. The biologist can select and explore suitable models in their graph representation and further evolve the domain theory as more knowledge becomes available.
Any hypothesis that should be tested can also be formalized as part of the domain theory.

\subsection{Specification using PALO}

Our formalization of the domain theory combines a generic theory of physical causality with axioms\footnote{Our theory of causality is loosely inspired by the theory of concurrency and causality that underlies Petri nets \cite{Stehr97} but greatly simplified.} taking into account observational evidence and some limited biological domain knowledge. The source of observational evidence is another neural network model that has been trained and validated to detect the existence of causality between genes solely based on their expression time profiles (modeled as Gaussian Processes) but without determining its direction. The details of this Siamese neural network model can be found in \cite{canes}, but are not essential for the following formalization,\footnote{A useful feature of Siamese networks is that they allow us to represent certain structural properties of relations, e.g., symmetry, directly in terms of the network structure. This is part of a general symmetry theme in our (equational) logic-based view discussed in Section \ref{sect-ext}.} which can be employed as long as we can obtain approximate probabilities for casual relations to start with.\footnote{In the RTA project we also developed models to predict direction, but we are intentionally using the simpler model here to avoid introducing additional uncertainty. An extension of this approach which incorporates both undirected and directed probabilistic causality is possible and has recently been implemented in the RTA workflow as well.} Utilizing a synthetic gene expression model, it is trained to determine the probability of a causal relationship between any two genes. The same model can be used to approximate independence, which we define as the probability of a causal relation being low.

As a basis we use a gene expression data set obtained by treating human cells with what was called $Unk5$ during a DARPA challenge \cite{Vertes18} and later revealed to be a common drug (atorvastatin) that regulates cholesterol biosynthesis.  We use PCA-based dimensionality reduction to generate an embedding of all protein-coding genes with significant perturbations (comparing treated against control timeseries), for which interesting causal relationships can be expected, in a 10-dimensional Euclidian space.\footnote{We use 10 for our gene dimension specification and 100 as a universal complexity specification (defining the family of learnable predicates). These hyperparameters were experimentally determined and reflect a tradeoff between computational resources and modeling precision.} The full space is represented by a type $Gene$ of dimensionality 10, and the sort $gene$ is bound to the relevant data set, namely the finite subset of all protein-coding genes with significant changes (approx. $1.2K$ out of $18K$ protein-coding genes).

One important application-dependent choice is the definition of the batch cover $\Delta$ used
to limit the batch size in the sampling semantics of quantifiers. Here we used a rough correlation-based analysis to establish that the effective sample size is larger than $50$, which should then be a reasonable choice for the batch size to maintain independence between genes in a single batch.

Two other sorts are needed to establish the link to experimental data: $lidata$ contains all pairs of genes for which causality can be detected with high probability (e.g., $\geq 0.7$ yielding $\approx 200K$ pairs) and $codata$ contains all pairs for which absence of causality is detected with high probability (e.g., $\geq 0.7$ yielding $\approx 650K$ pairs). It should be noted that in RTA, the data sets and most of these constants are determined by parameters, and the analysis is part of a larger workflow, but we are tying to keep things simple here to convey the basic idea.

Our domain theory is formalized by four binary relations, $li$, $co$, $di$, and $im$ (which can all be visualized as graphs),
which means they are of type $Gene\ Gene \rightarrow \prop$. The relations $li$ and $co$ stand for \textit{undirected causality} and \textit{independence} (concurrency). The relation $di$ stands for \textit{directed causality} and $im$ for \textit{immediate causality} (also directed). In the following we list and briefly motivate the axioms of the domain theory.

\paragraph{Basic axioms:}
$$ \forall i : gene\, . \, \neg li(i,i)
$$
$$ \forall i : gene \, . \, co(i,i)
$$
$$ \forall i,j : gene \, . \, li(i,j) \Rightarrow li(j,i)
$$
$$ \forall i,j : gene \, . \, co(i,j) \Rightarrow co(j,i)
$$
$$ \forall i,j : gene \, . \, li(i,j) \Rightarrow \neg co(i,j)
$$
$$ \forall i,j : gene \, . \, co(i,j) \Rightarrow \neg li(i,j)
$$
$$ \forall i,j : gene \, . \, co(i,j) \lor li(i,j) $$

\noindent
The basic axioms formalize irreflexitity of causality and reflexivity of independence (a useful convention
although other formalizations are possible). The two symmetry axioms reflect the undirected nature of these basic relations,
and the last two axioms express that these concepts are mutually exclusive and complementary.

\paragraph{Consistency with experimental data:}
$$ \forall p : lidata\, . \, li(p)
\quad \approx \quad 0.65 \ldots 0.75 $$
$$ \forall p : codata\, . \, co(p)
\quad \approx \quad 0.45 \ldots 0.95 $$

\noindent
Here we can interpret the geometric mean semantics of the universal quantifiers roughly as an approximation of
a minimum probability that is robust to outliers. While the lower bound on the probability predicted by our causality detector model for both causally dependent and independent pairs is $0.7$, we use the intervals $0.65 \ldots 0.75$ (that is $0.7 \pm 0.05$) and $0.45 \ldots 0.95$ (that is $0.7 \pm 0.25$), respectively, to account for some uncertainty (using a larger interval for independence due to the increased chance that long range causality can be mistaken as independence).\footnote{Our validation studies indicated that the causality detector model tends to slightly underestimate higher probabilities, but there are also some biological assumptions underlying that model that are reflected by our relatively large uncertainty intervals.}

\paragraph{Axioms for directed causality:}
$$ \forall i,j : gene \, . \, di(i,j) \Rightarrow li(i,j) 
$$
$$ \forall i,j : gene \, . \, li(i,j) \Rightarrow (di(i,j) \lor di(j,i))
$$
$$ \forall i,j : gene \, . \, di(i,j) \Rightarrow \neg di(j,i)
$$
$$ \forall i : gene \, . \, \neg di(i,i) $$
$$ \forall i,j,k : gene \, . \, di(i,j) \land di(j,k) \Rightarrow di(i,k)
$$

\noindent
Directed causality is formalized as a partial order that implies and generates undirected causality 
(first and second axiom). The subsequent axioms are simply the strict partial order axioms (irreflexivity, asymmetry,
and transitivity). Note that asymmetry is an example of an axiom that can be classically derived,
but it is stated explicitly due to its importance. The partial order formalizes global consistency of the
causal direction in the larger context but does not introduce a directional bias (that is at least two directions are
possible as in microscopic physics). Also it can be easily verified that the remaining axioms maintain this time reversal invariance.

\paragraph{Axioms for immediate causality:}
$$ \forall i : gene \, . \, \neg im(i,i)
$$
$$ \forall i,j : gene \, . \, im(i,j) \Rightarrow \neg im(j,i)
$$
$$ \forall i,j : gene \, . \, im(i,j) \Rightarrow di(i,j)
$$
$$ \forall i,k : gene \, . \, di(i,k) \land (\neg \exists j : gene \, . \, di(i,j) \land di(j,k)) \Rightarrow im(i,k)
$$
$$ \forall i,j,k : gene \, . \, im(i,j) \land im(j,k) \Rightarrow \neg im(i,k)
$$
$$ \forall i,j,k : gene \, . \, im(i,k) \land im(j,k) \Rightarrow co(i,j)
$$
$$ \forall i,j,k : gene \, . \, im(k,i) \land im(k,j) \Rightarrow co(i,j)
$$
$$ \forall i,j,k : gene \, . \, im(i,j) \land im(j,k) \Rightarrow li(i,k)
$$

\noindent
Immediate causality is a subset of directed causality (third axiom) such that an intermediate causal element does not
exist (fourth axiom). The other axioms are key properties that can be classically derived (theorems). They
are intended to make the soft-logic theory more precise and the approximate computational inference more efficient.
In effect, the last three are local consistency rules for causality and independence.

\paragraph{Estimates about density and degree of causality:}
$$ \mn{i,j : gene}{im(i,j)} \quad \approx \quad 0.001 \ldots 0.005
$$
$$ \forall j : gene \, . \, \neg \mn{i : gene}{im(i,j)} \quad \approx \quad 0.995 \ldots 1.0 $$
$$ \forall i : gene \, . \, \neg \mn{j : gene}{im(i,j)} \quad \approx \quad 0.995 \ldots 1.0 $$

\noindent 
In addition to the basic physical and experimental data axioms for causality, we can use some domain knowledge
to further narrow down the biologically plausible models. For instance, from curated gene expression networks we can
use estimates for the expected density of immediate causality (first axiom) and for the mean in- and out-degree
of typical networks (last two axioms), which is low, and most likely much less than $5$ out of $1000$ 
due to the approximately scale-free nature of these graphs\footnote{There is some experimental evidence for an
asymmetry between in- and out-degree that could be reflected by using more precise intervals.}. 
These examples also show how the mean probability quantifier can serve a useful purpose. 

Our probability intervals for experimental data and domain expertise are fairly wide to avoid inadvertently excluding feasible models, but it turns out that even such wide intervals are sufficient to narrow down the set of the most likely
models to a plausible subset (often with narrow ranges on the questionable parameters thanks to the multitude of other constraints and the large amount of data) that can be further analyzed quantitatively and inspected by a biologist.

\subsection{Sample Results}

\begin{figure*}[!htb]
	\includegraphics[width=1.0\linewidth]{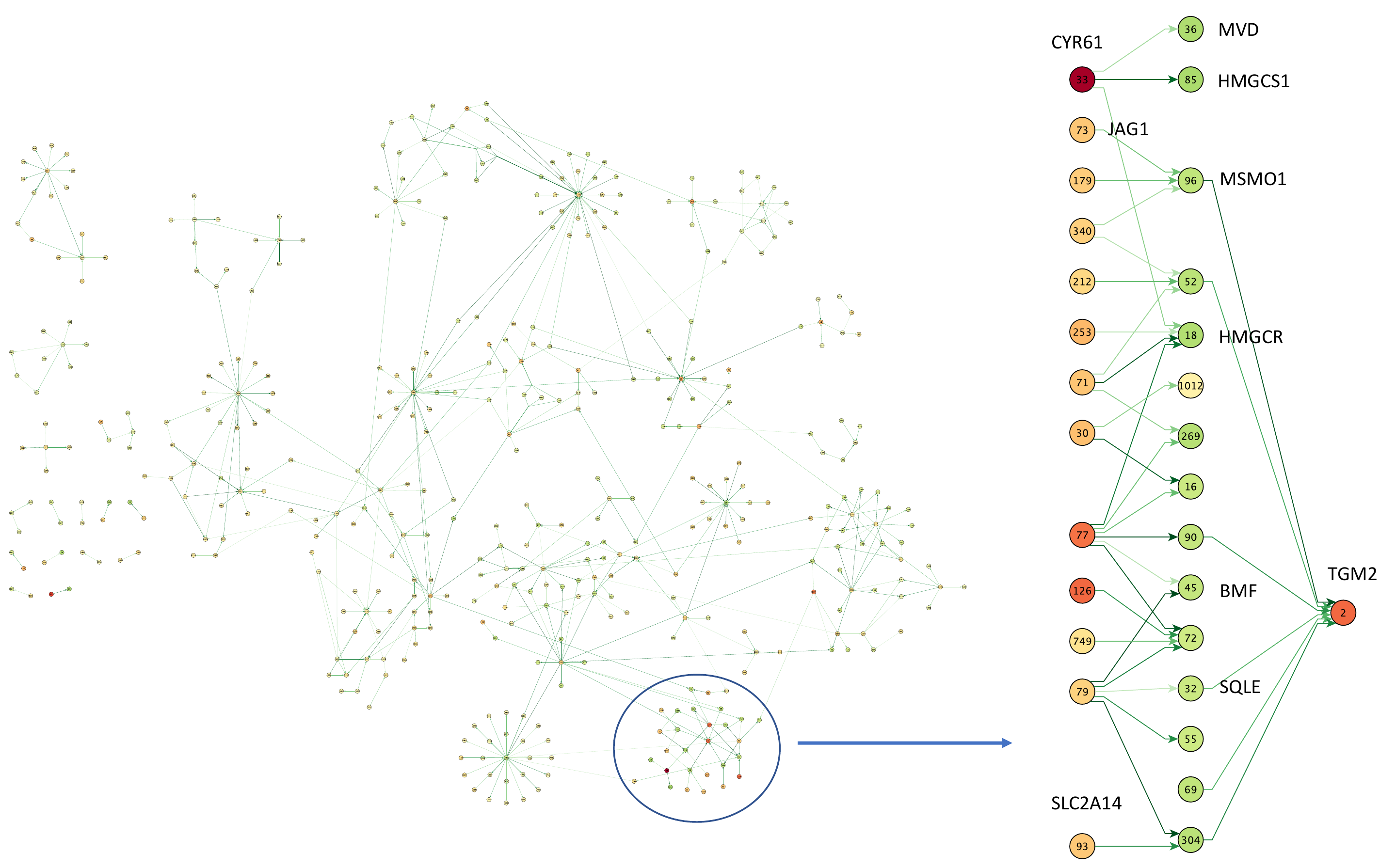}
	\caption{Sample Model with a Likelihood of 0.97. The subgraph on the right (related to the cholesterol pathway) is shown in context of the larger network limited to causal dependencies with probability at least $0.7$. Edges are colored so that darker colors correspond to higher probabilities. The node coloring reflects the average fold-change over the entire time series (green and red correspond to up and down regulation, respectively).}
	\vspace{5pt}
	\label{fig-pic}
\end{figure*}

In the following we visualize a sample model as a graph that depicts $im$, the immediate causality relation that can be extracted from the model using the model evaluation functionality described in the Section \ref{sect-lime}. The graph has been simplified by removing edges with a probability below $0.7$ and by removing isolated nodes, that is genes that do not exhibit any highly probable immediate causal connections. An automatically generated layout of the resulting graph with a subgraph that contains genes relevant to the mechanism of action of the drug is shown in Fig. \ref{fig-pic}. 

\begin{figure*}[!htb]
	\includegraphics[width=1.0\linewidth]{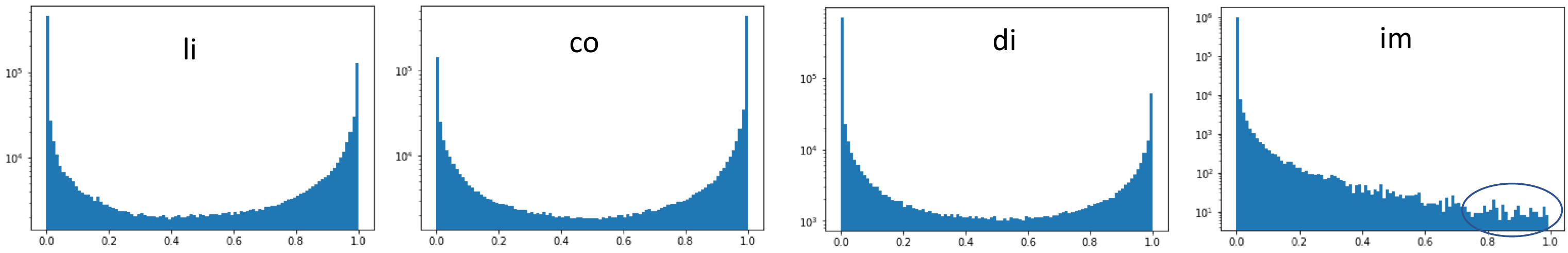}
	\caption{Relational Histograms of our Sample Model. Through sampling we can visualize the "probabilistic shape" of relations in our model. On the horizontal axis we show the probability, and on the logarithmic (!) vertical axis we show the number of samples (pairs of genes in the relation) with approximately that probability. The graph of Fig. \ref{fig-pic} roughly corresponds to the
		highlighted area in the $im$ relation.}
	\vspace{5pt}
	\label{fig-hist}
\end{figure*}

It is important to understand that this model represents one sample model in the posterior mode semantics, which is biased towards models with (locally) high likelihood. To understand the broader range of possibilities, for each instantiation of parameters, we typically generate 100 sample models in our automatic parallelized workflow on a cluster of GPU servers. In contrast to traditional deep learning, where a single model is usually sufficient and it has been argued that the non-convexity does not matter, our model space is strongly non-convex in a way that (partly) matters for the result. For instance, as already indicated above for each model of our theory the inverse model is equally likely, always leading to a complementary mode in the distribution.

The graphs underlying $li$, $co$, and $di$ are too large to show, but using model evaluation together with sampling
we can show abstractions, such as the histograms in Fig. \ref{fig-hist}, which can often shed light on the convergence of the model synthesis process. Comparing the histograms for $li$ and $co$, we can clearly see how their complementary nature also shows up at the statistical level. Furthermore, the fact that $li$ is approximately the symmetric closure of $di$ is consistent with the similarity in the shape of their histograms. Finally, $im$ is a very sparse subrelation of $di$ and $li$, which manifests itself in a highly asymmetric shape. 

\begin{figure*}[!htb]
	\includegraphics[width=1.0\linewidth]{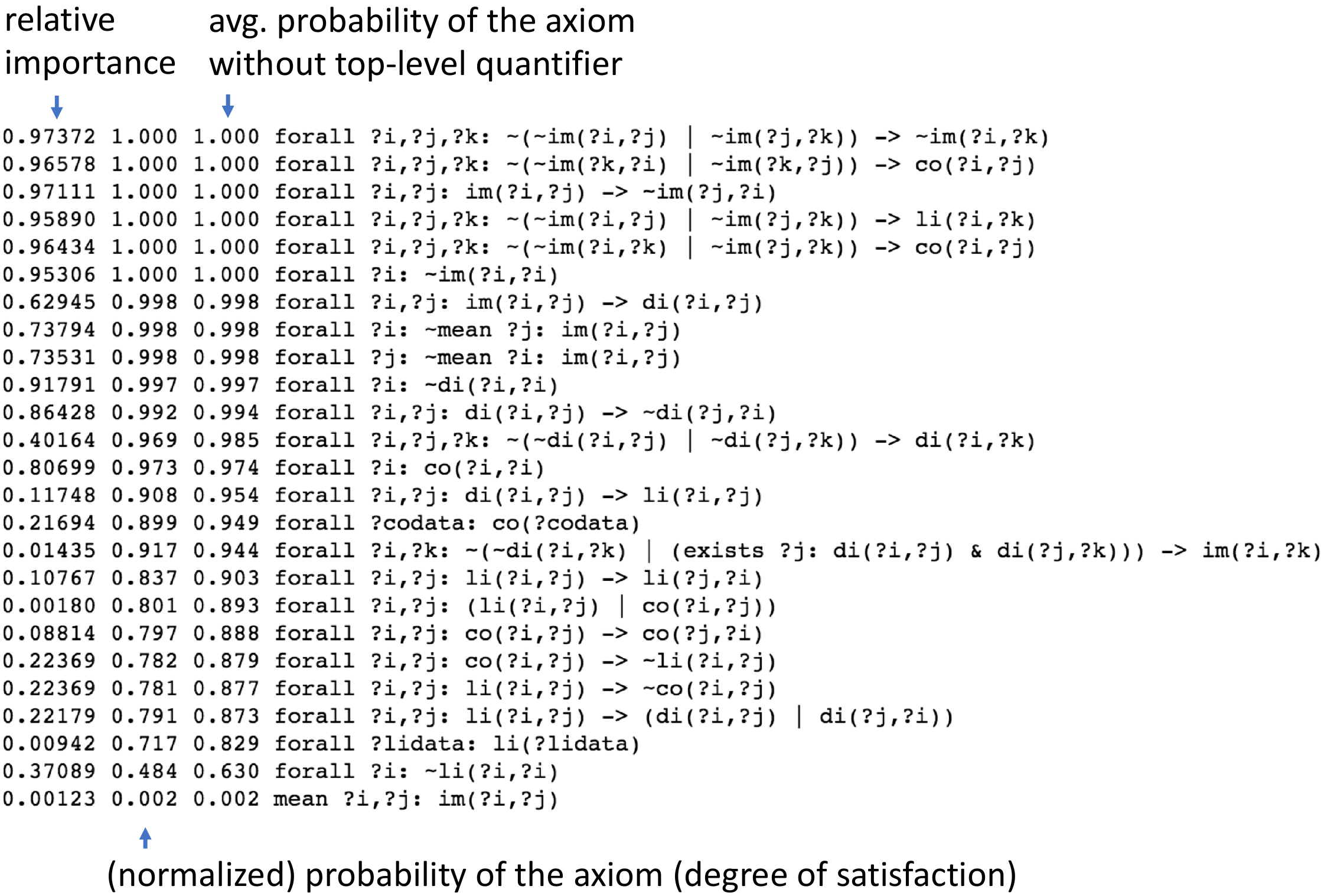}
	\caption{Mean Probability Analysis of the Sample Model. Using a slightly different logical form and notation, for each axiom we show the computed importance weight, the (normalized) probability, and the mean probability of the axiom without the top-level quantifier (which can be more intuitive for the user).}
	\vspace{5pt}
	\label{fig-res1}
\end{figure*}

While we used the model evaluation functionality to obtain Fig. \ref{fig-pic} we now use the model validation functionality to quantify the mean probability of the properties of interest. For sake of brevity, we focus on the axioms of our theory, even if in practice we may verify other implied and non-implied properties using the same functionality. In Fig. \ref{fig-res1} we
list each validated property together with three numbers. The first is the relative importance (that we also called the flexible weight) for each axiom. Recall that these are theory parameters that have been inferred during model synthesis together with all model parameters. The second number is the mean probability of the quantified property. The last number is the mean probability of the formula under the quantifier (in other words, the top-level quantifier is replaced by a mean quantifier), which is often more intuitive for the user. The validated properties
are ranked by their mean probabilities.

An interesting observation is that the importance/weight is not a very intuitive number and hence not necessarily
a parameter that should be exposed to the normal user. For example, the axiom involving $lidata$ is satisfied in the
model with an adequate (normalized) probability of more than $0.7$ (quite consistent with the capability of our causality detector which claims a lower bound of $0.7$), but the weight is relatively very low, which intuitively might suggest that
the axiom has not been heavily used to achieve/maintain this result (presumably partly because we have another axiom
involving $codata$ that is not independent and there are other axioms to infer undirected causality). A similar observation holds for the axiom with the existential quantifier that helps to generate immediate causality $im$. It should be noted, however, that in the PALO semantics, implication does not have a preferred direction so that any axiom involving $im$ can potentially contribute to the generation of new probabilistic pairs. Apart from this omnidirectional inference,
another factor that complicates the understanding of the approximate reasoning that takes place during model
synthesis is that by being intertwined with learning the notions of generalization and reasoning by similarity
have a major impact on the result and on how the axioms are used. For example a property established for one
gene may automatically transfer to similar genes, albeit with varying degree.

\begin{figure*}[!htb]
	\includegraphics[width=1.0\linewidth]{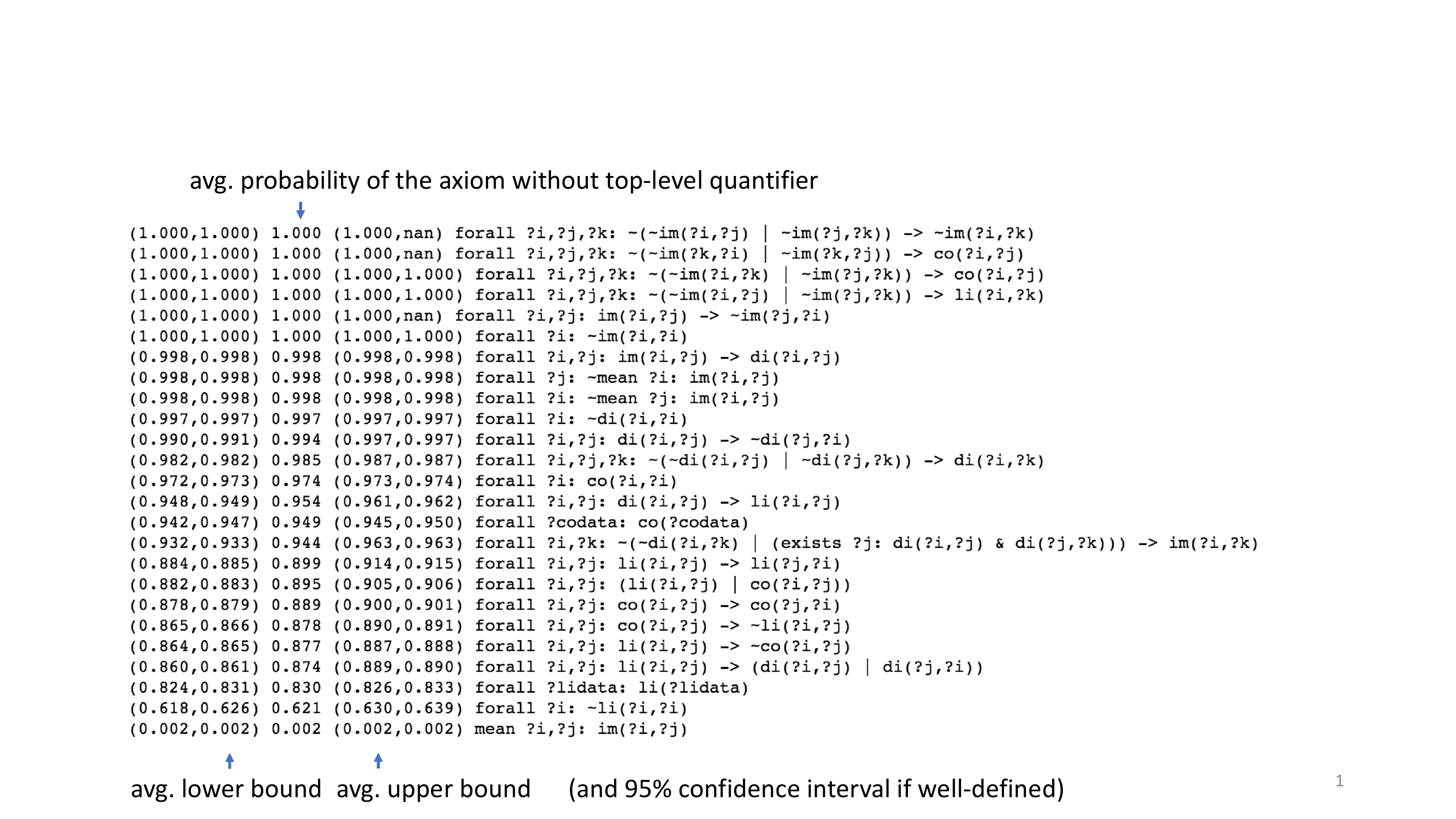}
	\caption{Lower and Upper Bound Mean Probability Analysis of the Sample Model. We show only the mean probabilities
		of each axiom without the top-level quantifier and the corresponding lower and upper bounds with confidence intervals.}
	\vspace{5pt}
	\label{fig-res2}
\end{figure*}

Finally, we would like to gain an understanding of the precision of our approximate validation for the given model. One might 
expect a high imprecision as the approximate probability does not account for logical dependencies. To this end, we use
the full semantics that includes lower and upper mean probabilities. The results of model validation using this
semantics are shown in Fig. \ref{fig-res2}. For a better intuition, we validate the mean probabilities of all axioms
without the universal quantifier. It turns out that even we add $95\%$ confidence intervals the bounds
are very tight in spite of the fact that we only used 100 samples of sort-bindings (of modest batches of 
size $50$ for genes and $250$ for pairs) to compute the mean. This shows that our data sets contain 
enough diversity and our theory is suitably structured to achieve a very good precision in the relevant 
mean probabilities.

\begin{figure*}[!htb]
	\includegraphics[width=1.0\linewidth]{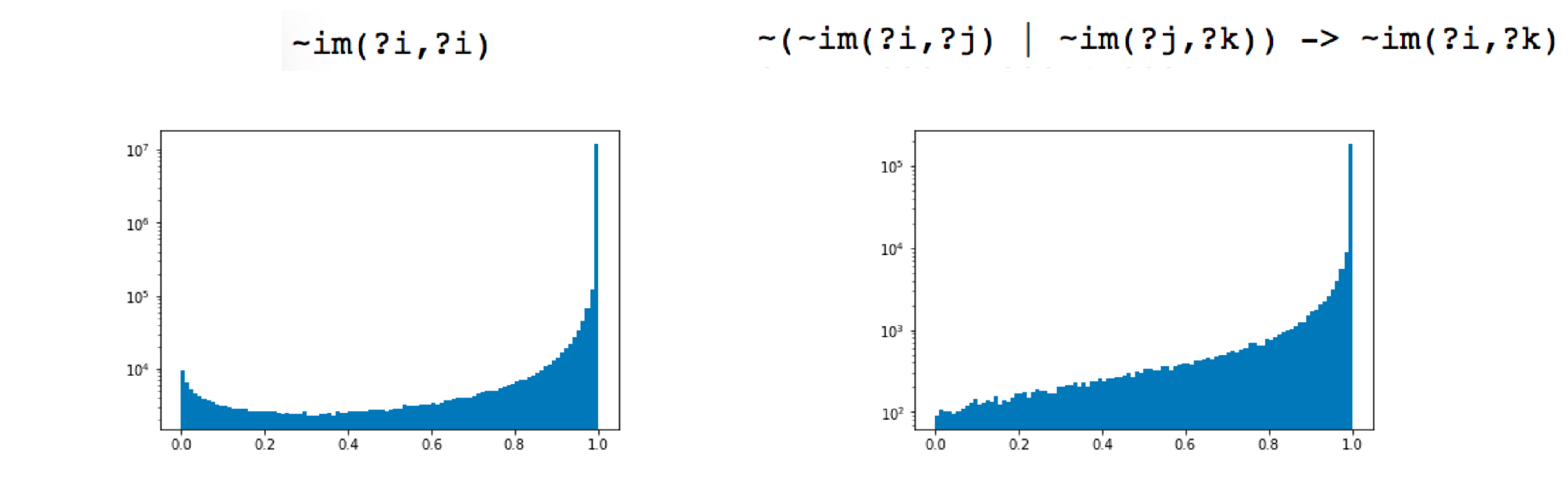}
	\caption{Propositional Histograms for our Sample Model. Through sampling we can also visualize the "probabilistic shape" of the axioms (and other properties) in our model. To this end, we again consider the axiom without the top-level quantifier. Two examples are given for axioms that seem to exhibit a perfect mean probability of $1.0$, but it is important to note that the mean probability can still hide their detailed characteristics. The logarithmic scale is essential to make such imperfections visible. }
	\vspace{5pt}
	\label{fig-hist2}
\end{figure*}

It is noteworthy that sampling-based model evaluation can be not only used to extract relations, but it can be applied 
to any term in the logic, in particular to propositional terms, that is formulas. This can be used to obtain more details about the satisfaction of an axiom or any other property in a given model as illustrated in Fig. \ref{fig-hist2}.

We have illustrated approximate computational inference using the primary semantics of PALO, namely the mean probability semantics with lower and upper bounds, but we like to point out that the results are based on a formalization of the domain theory that is sufficiently explicit and hence computationally efficient for our purposes. Deriving such richer theories from a
small set of basic axioms is an interesting topic by itself, and a place where the abstract classical semantics is essential. Our logical theory was simple enough to verify the derived axioms manually under this semantics, but more complex domain theories may benefit from automatic symbolic inference (see Section \ref{sect-ext} on possible extensions of LIME). 

\begin{figure*}[!htb]
	\includegraphics[width=1.0\linewidth]{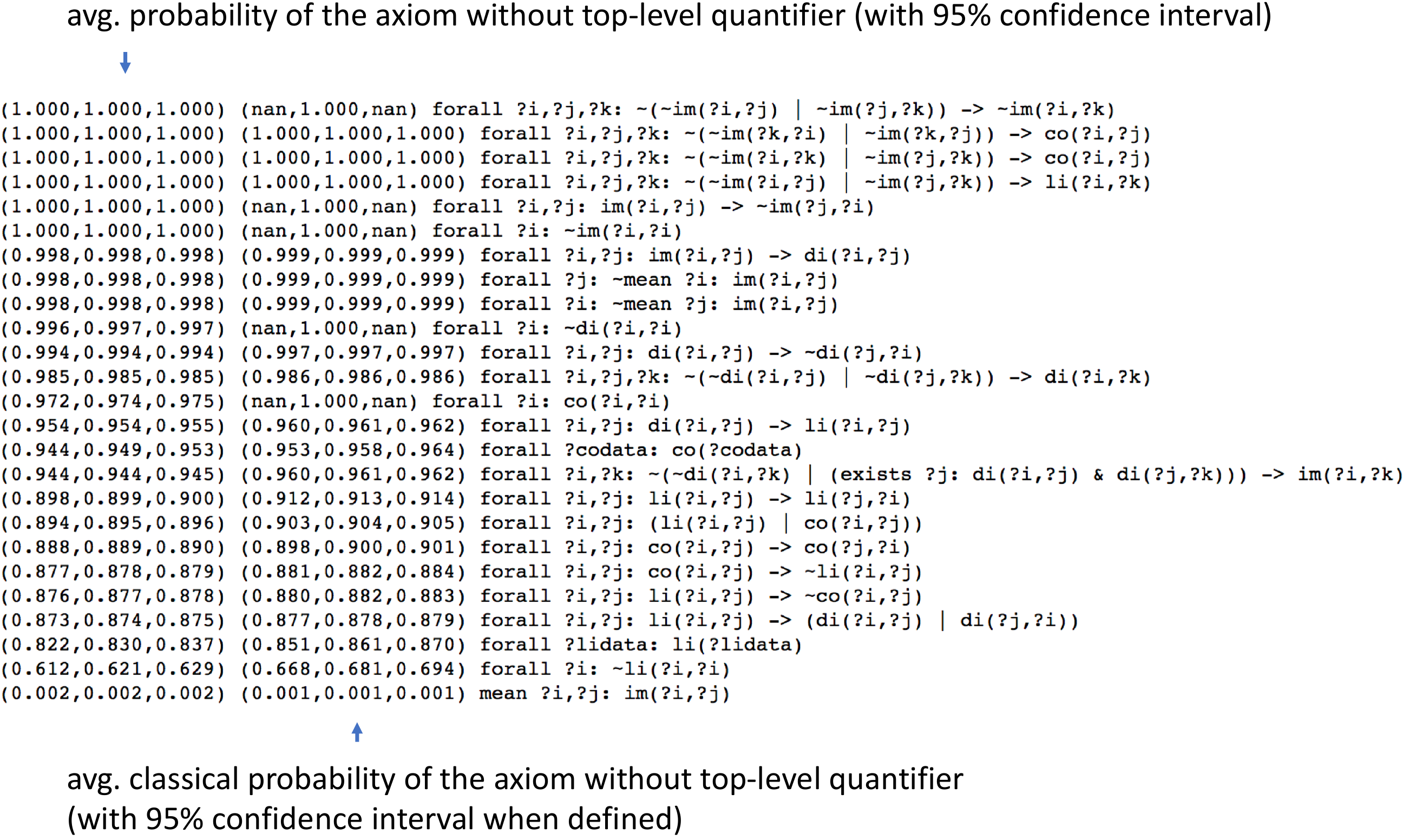}
	\caption{Comparing Mean Probabilities under the Approximate (Soft) Semantics and under the Concrete Classical Semantics. In spite of the fact, that crispification leads to a significant loss of information for individual instantiations
	of formulas, it turns out that the mean probabilities are quite similar in our application (albeit with clearly noticeable differences). }
	\vspace{5pt}
	\label{fig-res3}
\end{figure*}

Finally, the concrete classical semantics can be used to evaluate the imprecision introduced by our soft logic approximation, against a classical semantics which necessarily suffers from a very different type of imprecision caused by crispification. The results are shown in Fig. \ref{fig-res3} and are quite acceptable for this type of application that involves many other sources of uncertainty.

We conclude this section with another caveat regarding our formalization. Immediate causality, which is the basis for our biological network graphs, is relative to the chosen level of abstraction, which is a subset of genes in our example. The reality is far more complex, as some protein-coding genes encode transcription factors, that is proteins that again regulate gene expression in the context of other transcription factors in a complex fashion that can favor up or down regulation. More complete networks with proteins and positive and negative dependencies have been studied in the RTA project as well. We do not expect that applying PALO and LIME to such networks would require fundamental changes in the theory. On the other hand, a logical treatment of our more abstract cluster-based graph synthesis \cite{Vertes18} would lead to some modifications and could be an interesting topic for future work.

The time reversal invariance of our theory means that if one globally consistent model exists, a model with inverted $di$ and $im$ relations is equally probable. This is a consequence of not using any directional causality as input. What might be surprising is that in practice the networks (that is the graph generated by $im$) are sufficiently connected that only two possible models remain, and if the direction of causality is fixed anywhere in the network, the direction is globally uniquely
determined. Hence, it is easy for a biologist to select the proper time direction even without using
other directional models developed in the RTA project. This is, however, not necessary with our most recent addition
to the RTA workflow, where through an extension of the logical theory presented here we accommodate knowledge about causality and its direction (each with its own degree of uncertainty) effectively using PALO to consistently integrate both of the Siamese neural network models presented in \cite{canes}.

While the properties of our synthesized models can be objectively validated (in particular independent of the flexible
weights that are only relevant during model synthesis and might better be hidden), it may come as no surprise that the model synthesis process is not fully understood in detail. This seems similar to the problem of understanding deep learning processes, where the high dimensionality of the model space leads to counterintuitive properties. The additional difficulty with logical theories is that the model space is usually multi-modal in an essential way, which further complicates the matter. Overall we expect that  many more case studies are needed to get a better intuition for the characteristics of approximate computational inference that happens during model construction.

\section{Extensions of the Core Logic and its Engine}\label{sect-ext}

To evolve PALO into a practical general-purpose language for machine learning applications a lot of work remains to be done. Some features such as built-ins and semantic attachments are straightforward to add and facilitated by using the generalized LTN framework. Other more fundamental features, such as equations require careful experimentation to understand the computational feasibility and quality of the approximation. In the following, we list the in our view most interesting directions for future extensions and generalizations ranked by increasing difficulty and level of effort.

\paragraph{Built-ins, Semantic Attachments, and Richer Models}

Although not necessary for our bioinformatics application, most applications have special needs
for arithmetic and logical\footnote{There is no need for new logical operators in classical logics, but as a soft logic PALO is incomplete w.r.t.\ a classical semantics. Hence additional operators can increase expressiveness.} operators, and we have already indicated in our presentation of PALO, where they can be added as built-in symbols (with fixed interpretations). In our prototype such operators could be defined in TensorFlow/Python and are equivalent to semantic attachments in
conventional logics. Our tensor abstractions ($\lambda$) provide a convenient way to also introduce new higher-order operators like min and max operators for arbitrary terms. 

A more general notion of semantic attachment (with a flexible interpretation that depends on new model parameters) can be introduced by associating predicate and function symbols with other learnable families of semantic functions that can be represented as TensorFlow graphs with variables (e.g., deep neural networks as black-box components). To this end, suitable types (and extended complexity specifications if needed) should be introduced for each new family to keep the interpretation separate from the existing predicate and functions symbols.
For example, in addition to the set of linear functions (also used in Real Logic and LTNs to interpret function symbols)
other classes of functions of practical interest can be added such as multivariate polynomials and subclasses, e.g., the class of functions used in factorization machines \cite{Rendle10}. Richer interpretations for predicates are also conceivable, for example their representation as deep Sum-Product Networks \cite{Poon11} or as parameterized families of domain/application-specific neural network architectures such as deep convolutional networks. To reduce complexity and improve scalability, learning could proceed hierarchically, which is connected to the topic of compositionality and modularity (see below).

\paragraph{Equational Logic and Structural Symmetries}

An important concept that we envision to add to PALO is approximate equality. The power of equality is witnessed by many of the traditional symbolic logical systems. For example, in Maude \cite{maude} which is based on \textit{membership equational logic} \cite{meq-logic}, a conditional equational logic with sorts, the focus on equality leads to very intuitive executable specifications and supports its use as a logical and semantic framework for other formalisms \cite{rwl-as-framework}. Clearly, without equality the use of functions is somewhat limited. Interestingly, there is not a single concept of equality once we accept that it must be approximate. A symmetric definition using Euclidean distance is only one possibility,\footnote{See also \cite{Gaines78} for his definition of logical equivalence in terms of logical distance that is consistent with our definition in PALO.} and there is also an alternative view suggested by \textit{rewriting logic} \cite{cond-rw-logic}, which itself was inspired by linear logic and can be regarded as an extreme substructural logic where even the symmetry of equality is given up. This leads to a directed interpretation of equality in terms of rewriting with many interesting logical and semantic framework applications such as in the representation of deductive or concurrent systems. Approximating this notion would lead to a more refined notion of equality. We envision that like in Maude both notions should be able to peacefully coexist. There is also a third notion of \textit{structural equality} in Maude, that is based on built-in equational theories. Such structural equations can also be a powerful feature as they may allow us to express \textit{symmetries} that can be efficiently realized by submodules such as deep symmetry networks \cite{deep-symmetry}.
Structural equality can be further generalized to (logical) \textit{structural equivalence}. For example, it is conceivable that
certain common properties of predicates (e.g., reflexivity, irreflexivity, symmetry, asymmetry, transitivity) can be
enforced structurally by suitable classes of neural networks (the Siamese networks used in \cite{canes} are only one example) resulting in increased learning and inference efficiency. This naturally leads to the topic of modularity of logical theories and their corresponding neural networks.

\paragraph{Modularity and Composability of Theories}

To cover this topic it is useful to adopt a \textit{dual view of logic} as a means of knowledge representation as well as an architectural description language. It suggests that the natural modular structure of the domain theory (including any inferred domain knowledge) should lead to a corresponding modular structure of the neural network that is synthesized from it. The modular structure of the neural network (which can be made explicit using constructs such as hierarchical scoping in TensorFlow \cite{tensorflow}) can then be exploited for compositional training strategies. For example, a basic strategy of \textit{concurrent training} simply trains components independently and composes them afterwards. In \textit{sequential training}, components are trained layerwise, by training and freezing the models of lower layers before higher layers are trained. More flexible strategies perform \textit{pretraining} \cite{deep-learning} of basic components before they are composed and then continue the training process in a larger context without freezing the pretrained models, thus allowing the smaller modules adapt to the bigger context. 

To summarize, conventional approaches to deep learning do not have access to an \textit{explicit representation of the domain theory}, which is a big disadvantage that must be compensated by manual, error-prone engineering. Thanks to our logical framework, we can exploit the structure of the theory to automatically synthesize the proper architecture of the neural network and corresponding training strategies tailored to the network structure, the type of problem, and the type of data. Another practical benefit of our envisioned modular approach is the ability to integrate existing well-tested \textit{black-box components} (i.e., modules that are not necessarily synthesized from a logical theory), such as highly-optimized and well-tested convolutional network architectures for image recognition \cite{deep-learning}. In such cases, the domain theory can refer to the interfaces of such components (e.g., detectable features) without referring to their internal structure.

\paragraph{Towards a Neural-Symbolic Architecture}

Machine learning and symbolic reasoning technologies have been evolving rapidly in the last decade, albeit mostly independently and driven by very different technologies. Deep learning architectures have been enabled by new, modular approaches to training large neural networks, and they are simultaneously exploiting the explicit representation of dataflows for efficient mappings into specialized, highly parallel hardware and reflective metalevel algorithms (e.g., automatic differentiation). Symbolic reasoning engines, on the other hand, can now can perform millions of inferences per second and deal with problems involving millions of variables, often enabled by highly optimized data structures and algorithms that make efficient use of general purpose CPU architectures.

With its dual approximate and classical semantics, PALO may offer a possible semantic foundation for the synergistic integration of these emerging technologies. Specifically, we envision an application-independent \textit{neural-symbolic architecture} that supports dataflows in which \textit{approximate computational inference} and \textit{exact symbolic inference} can take place in a loosely coupled fashion with \textit{bidirectional knowledge transfer} between symbolic and neural forms of representation. We think a lose coupling, most suitably in the form of a distributed and scalable networked implementation, is essential to maintain the high efficiencies of the participating technologies. In such a framework, key research questions will be concerned with the types of knowledge transfers and the distributed execution strategy. Ideally, a well designed general purpose architecture should support multiple workflows including automated ones such as model learning with subsequent validation as well as interactive workflows, e.g., knowledge engineering and model discovery with interactive theory refinement.

\section{Conclusion}\label{sect-concl}

We have focused on the use of logic as a systematic means of incorporating domain knowledge into machine learning
to address a key conceptual and engineering roadblock that is becoming increasingly important with the \textit{growing diversity} of deep learning applications. Here we would like to point out that there are other \textit{critical limitations} of deep learning architectures, where we expect that a logic-based approach can lead to improvements. 
For example, our approach should reduce the need for large annotated training sets by placing domain knowledge on the same footing as and using it as a substitute for empirical data (supporting semi-supervised and even unsupervised learning).
An equational extension of PALO, can furthermore be used to express feature space symmetries, which can drastically reduce the amount of training data. The limitation of current approaches that at best can only perform shallow inference is partly addressed by PALO and LIME, but would be more completely addressed by extending LIME to a neural-symbolic architecture that can concurrently operate on neural representations optimized for (deep) learning and symbolic representations that can enable (deep) inference. Finally, we should mention the increasing range of vulnerabilities of deep learning systems, e.g., safety concerns highlighted by adversarial examples or privacy issues caused by learned representations that are too informative. While there are partial generic solutions, it is unlikely that such problems can be fully solved in a way that is completely agnostic to the application and the system context, hence domain knowledge, logical structure, inference, and a generalized notion of validation, can be expected to play a more important role in these areas as well.

In PALO, that is heavily based on recent advancements in machine learning and neural networks, 
data and knowledge are treated on an equal footing. Both are incorporated uniformly using axioms of the
underlying theory and usually exhibit uncertainly that can be approximately and probabilistically quantified.
A theory is specified by a finite (often incomplete) set of relevant quantifiable axioms that are expected properties of our model. Such axioms might also be considered direct observables of our models, and the role of our logical imagination engine is to synthesize models only based on these directly observable properties. Although a better understanding of the approximate inference process would be valuable, through an independent validation we can assure not only that these axioms hold and to what degree, but also validate and quantify any potentially implied properties. We may also identify models with unexpected properties that may lead to new discoveries, hypothesis and refined theories in this knowledge engineering and model synthesis/validation cycle. 

More generally, the role of a logical theory is that of a domain-specific regularizer for the set of models, which can partly compensate for a lower amount or quality of data, and which can
constrain the underlying neural networks to maintain a whole range of properties with corresponding tradeoffs.
Quite different from traditional symbolic inference systems or even model checkers, most computational resources
are spent on the synthesis of models (this is where the use of highly parallel hardware such as GPUs is essential), as opposed to the validation of their properties. The validation can be highly efficient (even without GPUs), which means that we can think of a model as a highly compressed joint representation of theory and data in a form that can be efficiently queried, potentially leading to new logical applications, e.g., on mobile devices.

Finally, we would also like to recall from our earlier discussion that there is a complementary viewpoint to the role of PALO as a language for knowledge representation. We may equally think of it as the core of an architectural description language for machine learning architectures. From a structural point of view, a logical theory formalizes how black- or white-box components are combined to achieve an overall performance objective (measured in terms of the observable properties). This view provides an alternative motivation for several of our proposed extensions, for example the emphasis on equational logic to express symmetries that can be directly realized by suitable architectural families. This is very much in analogy to theories that are architecturally built into symbolic engines such as Maude \cite{maude} and Yices \cite{yices}, so that reasoning modulo such theories becomes highly efficient. Composability and modularity are other desirable features of both languages for knowledge representation and for architectural specifications, and hence important areas for future work.

\vspace*{1ex}


\noindent
{\bf Acknowledgements}
The reported research was sponsored by the U.S. Army Research Office and the Defense Advanced Research Projects Agency and was accomplished under Cooperative Agreement Number W911NF-14-2-0020. The views and conclusions contained in this document are those of the authors and should not be interpreted as representing the official policies, either expressed or implied, of the Army Research Office, DARPA, or the U.S. Government. The U.S. Government is authorized to reproduce and distribute reprints for Government purposes notwithstanding any copyright notation hereon. 

\vspace*{1ex}

\noindent
{\bf Author Contributions}
This paper is an initial attempt to nail down the theoretical foundations for the broader concept of
"Imagination Modulo Logical Theories" that Mark-Oliver Stehr developed in a white paper with support by Minyoung Kim and Carolyn Talcott. Mark-Oliver Stehr developed the theoretical foundations of PALO and the prototypical implementation of LIME. He wrote this paper with input from Carolyn Talcott, Minyoung Kim, and Merill Knapp. Additionally, Minyoung Kim contributed to the testing of our underlying framework. Finally, Merill Knapp, Carolyn Talcott, and Akos Vertes contributed their systems biology expertise to the interpretation and discussion of our results in the context of the RTA project.

\clearpage

\bibliographystyle{abbrv}
\bibliography{ref}

\end{document}